\documentclass[conference]{IEEEtran}
\IEEEoverridecommandlockouts
\usepackage[sort&compress,numbers]{natbib}%\usepackage{cite}
\usepackage{amsmath,amssymb,amsfonts}
\usepackage{algorithmic}
\usepackage{graphicx}
\usepackage{textcomp}
\usepackage{xcolor}
\def\BibTeX{{\rm B\kern-.05em{\sc i\kern-.025em b}\kern-.08em
    T\kern-.1667em\lower.7ex\hbox{E}\kern-.125emX}}

\makeatletter
\newcommand{\linebreakand}{%
  \end{@IEEEauthorhalign}
  \hfill\mbox{}\par
  \mbox{}\hfill\begin{@IEEEauthorhalign}
}
\makeatother

\usepackage{xspace}
\usepackage{hyperref}
\usepackage{listings}
\usepackage{enumitem}
\usepackage{multirow}
\usepackage{booktabs}
\usepackage{array}
\usepackage{subcaption}
\usepackage{pifont}
\usepackage{fancyvrb}
\usepackage{optidef}

\usepackage{orcidlink}
\usepackage{tcolorbox}

\usepackage[ruled,linesnumbered]{algorithm2e}

\graphicspath{{figures/}}

\usepackage[acronym]{glossaries}

\hypersetup{%
  colorlinks=true,
  linkcolor=black,
  citecolor=black,
  urlcolor=black
}

\newacronym{OWASP}{OWASP}{Open Web Application Security Project}
\newacronym{CRS}{CRS}{Core Rule Set}
\newacronym{regex}{regex}{regular expression}
\newacronym{WAF}{WAF}{Web Application Firewall}
\newacronym{ML}{ML}{machine learning}
\newacronym{RL}{RL}{Reinforcement Learning}
\newacronym{AI}{AI}{Artificial Intelligence}
\newacronym{NLP}{NLP}{Natural Language Processing}
\newacronym{SVM}{SVM}{Support Vector Machine}
\newacronym{LR}{LR}{Logistic Regression}
\newacronym{SecSVM}{SecSVM}{Secure Support Vector Machine}
\newacronym{RF}{RF}{Random Forest}
\newacronym{CNN}{CNN}{Convolutional Neural Networks}
\newacronym{SQLi}{SQLi}{SQL injection}
\newacronym{XSS}{XSS}{cross-site scripting}
\newacronym{RCE}{RCE}{remote code execution}
\newacronym{PL}{PL}{Paranoia Level}
\newacronym{TNR}{TNR}{True Negative Rate}
\newacronym{FPR}{FPR}{False Positive Rate}
\newacronym{TPR}{TPR}{True Positive Rate}
\newacronym{ROC}{ROC}{Receiver Operating Characteristic}
\newacronym{DR}{DR}{Detection Rate}
\newacronym{API}{API}{Application Programming Interface}

\graphicspath{{figures/}}

\newcolumntype{C}[1]{>{\centering\let\newline\\\arraybackslash\hspace{0pt}}m{#1}}

\makeatletter
\AtBeginDocument{\let\c@listing\c@lstlisting}
\makeatother

\begin{document}
\newcommand{\sota}{state-of-the-art\xspace}
\newcommand{\diff}[2]{\frac{\partial #1}{\partial #2}}
\newcommand{\vct}[1]{\ensuremath{\boldsymbol{#1}}}
\newcommand{\mat}[1]{\ensuremath{\mathbf{#1}}}
\newcommand{\set}[1]{\ensuremath{\mathcal{#1}}}
\newcommand{\con}[1]{\ensuremath{\mathsf{#1}}}
\newcommand{\tsum}{\ensuremath{\textstyle \sum}}
\newcommand{\T}{\ensuremath{\top}}
\newcommand{\mycomment}[1]{}
\newcommand{\lc}[1]{\mycomment{LC: #1}}
\newcommand{\bm}[1]{\mycomment{BM: #1}}
\newcommand{\ind}[1]{\ensuremath{\mathbbm 1_{#1}}}
\newcommand{\argmax}{\operatornamewithlimits{\arg\,\max}}
\newcommand{\erf}{\text{erf}}
\newcommand{\sign}{\text{sign}}
\newcommand{\argmin}{\operatornamewithlimits{\arg\,\min}}
\newcommand{\bmat}[1]{\begin{bmatrix}#1\end{bmatrix}}
\newcommand{\myparagraph}[1]{\noindent \textbf{#1}}
\newcommand{\myparagraphlb}[1]{\noindent \newline \textbf{#1}}
\newcommand{\mysubparagraph}[1]{\noindent \underline{\textit{#1}}}
\newcommand{\ie}{i.e.\xspace}
\newcommand{\eg}{e.g.\xspace}
\newcommand{\etc}{etc.\xspace}
\newcommand{\aka}{a.k.a.\xspace}
\newcommand{\wrt}{w.r.t.\xspace}
\newcommand{\etal}{\emph{et al.}\xspace}

\newcommand{\advms}{ModSec-AdvLearn\xspace}
\newcommand{\mlms}{ModSec-Learn\xspace}
\newcommand{\modsecurity}{ModSecurity\xspace}
\newcommand{\wafamole}{WAF-A-MoLE\xspace}
\newcommand{\sqligot}{SQLiGoT\xspace}
\newcommand{\modseclearn}{ModSec-Learn\xspace}

\renewcommand{\sectionautorefname}{Sect.}
\renewcommand{\subsectionautorefname}{Sect.}
\renewcommand{\figureautorefname}{Fig.}
\renewcommand{\equationautorefname}{Eq.}
\renewcommand{\algorithmautorefname}{Alg.}
\newcommand{\red}[1]{\textcolor{red}{#1}}

\title{
\advms: Countering Adversarial SQL Injections with Robust Machine Learning
}

% \author{Biagio~Montaruli~\orcidlink{0009-0002-6870-8075},
%         Giuseppe~Floris~\orcidlink{0009-0007-7000-5260},
%         Christian~Scano~\orcidlink{0009-0003-7756-6125},\\
%         Luca~Demetrio~\orcidlink{0000-0001-5104-1476},
%         Andrea~Valenza~\orcidlink{0000-0001-7771-2485},
%         Luca~Compagna~\orcidlink{0009-0003-1072-4352},
%         Davide~Ariu~\orcidlink{0000-0001-7970-5959},
%         Luca~Piras~\orcidlink{0000-0001-5135-0172},\\
%         Davide~Balzarotti~\orcidlink{0000-0001-5957-6213},
%         and~Battista~Biggio~\orcidlink{0000-0001-7752-509X},~\IEEEmembership{Senior~Member,~IEEE}% <-this % stops a space
\author{
        Giuseppe~Floris\textsuperscript{\textsection},
        Christian~Scano\textsuperscript{\textsection},
        Biagio~Montaruli\textsuperscript{\textsection},
        Luca~Demetrio\textsuperscript{*},
        Andrea~Valenza,\\
        Luca~Compagna,
        Davide~Ariu,
        Luca~Piras,
        Davide~Balzarotti,
    and~Battista~Biggio\textsuperscript{*},~\IEEEmembership{Fellow,~IEEE}% <-this % stops a space

\thanks{G. Floris, C. Scano and B. Biggio are with the Dept. of Electrical and Electronic Engineering, University of Cagliari, 09124 Cagliari, Italy e-mail: (name.surname@unica.it), C. Scano is also with the Department of Computer, Control and Management Engineering, Sapienza University, Rome, Italy.}%
\thanks{B. Montaruli and D. Balzarotti are with the Dept. of Digital Security, EURECOM, 06410 Biot, France, e-mail: (name.surname@eurecom.fr).}%
\thanks{Luca Demetrio is with the Department of Informatics, Bioengineering, Robotics and Systems Engineering (DIBRIS), University of Genova, 16146 Genova, Italy e-mail: (luca.demetrio@unige.it).}%
\thanks{Andrea Valenza is with Prima Assicurazioni, 20131 Milano, Italy e-mail: (andrea.valenza@prima.it).}%
\thanks{Luca Compagna is with Endor Labs, e-mail: (lcompagna@endor.ai).}%
\thanks{Davide Ariu and Luca Piras are with Pluribus One, 09128 Cagliari, Italy e-mail: (name.surname@pluribus-one.it).}
\thanks{\textsuperscript{\textsection} means equal contribution, while \textsuperscript{*} refers to corresponding authors.}}

\maketitle

\begin{abstract}
Many Web Application Firewalls (WAFs) leverage the OWASP \gls{CRS} to block incoming malicious requests. The CRS consists of different sets of rules designed by domain experts to detect well-known web attack patterns. Both the set of rules and the weights used to combine them are manually defined, yielding four different default configurations of the \gls{CRS}. In this work, we focus on the detection of \gls{SQLi} attacks, and show that the manual configurations of the \gls{CRS} typically yield a suboptimal trade-off between detection and false alarm rates. Furthermore, we show that these configurations are not robust to adversarial \gls{SQLi} attacks, \ie, carefully-crafted attacks that iteratively refine the malicious \gls{SQLi} payload by querying the target WAF to bypass detection. 
To overcome these limitations, we propose (i)~using machine learning to automate the selection of the set of rules to be combined along with their weights, \ie, customizing the \gls{CRS} configuration based on the monitored web services; and (ii)~leveraging adversarial training to significantly improve its robustness to adversarial \gls{SQLi} manipulations.
Our experiments, conducted using the well-known open-source ModSecurity WAF equipped with the CRS rules, show that our approach, named \advms, can (i)~increase the detection rate up to 30\%, while retaining negligible false alarm rates and discarding up to 50\% of the \gls{CRS} rules; and (ii)~improve robustness against adversarial \gls{SQLi} attacks up to 85\%, marking a significant stride toward designing more effective and robust WAFs. We release our open-source code at \url{https://github.com/pralab/modsec-advlearn}.
\end{abstract}


\begin{figure*}[!th]
    \centering
    \includegraphics[width=\textwidth]{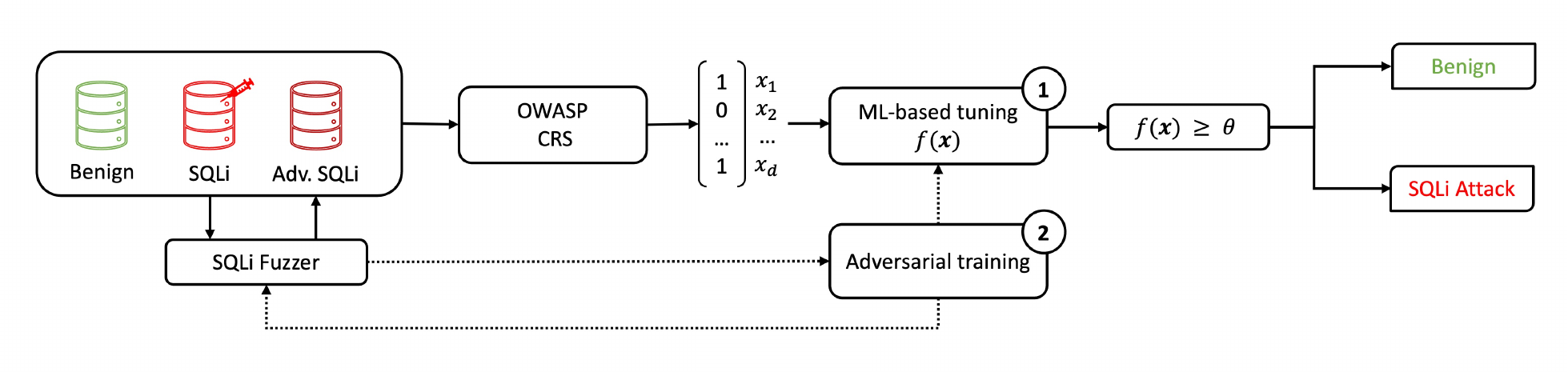}
    \caption{Conceptual representation of \advms. A machine-learning model is trained using the \gls{CRS} rules as input features, and leveraging our novel adversarial training approach to improve robustness against adversarial SQLi attacks.}
\label{fig:system_overview}
\end{figure*}

\section{Introduction}\label{sec:introduction}
Web applications are constantly evolving and deployed at a broad scale, thus enabling organizations to offer rich services over the Internet.
However, this imposes serious challenges in securing web applications against an increasing number of attacks \cite{web_sec_survey}.
Among these, \gls{SQLi} consists of injecting a malicious SQL code payload inside regular queries, causing the target web application to behave in an unintended way or expose sensitive data.
Even if many countermeasures to this attack have been proposed~\cite{amnesia2006sql,joshi2014sql,appelt2017tuningwaf,ml_driven}, the \gls{OWASP} Foundation still classifies it as one of the top-10 most dangerous web threats~\cite{owasp_top_ten}.

\glspl{WAF} are commonly used as a defense tool in enterprise systems to counter such attacks and protect web applications~\cite{waf_survey,ml_driven}.
They work by filtering the incoming requests directed towards the web applications and blocking suspicious connections. To this end, many available \gls{WAF} solutions leverage the OWASP \acrlong{CRS}, i.e., a collection of signatures designed to detect well-known web attack patterns. 
The \gls{CRS} rules have all been developed by experts in the domain of web security in the last decade, helping to withstand a vast plethora of web-based attacks. The \gls{CRS} v4.0.0 (one of the latest stable versions) used in this work includes 319 rules, out of which 170 target critical injection attacks~\cite{crs_doc}. Within this set, \gls{SQLi} is the most represented class of injection attack counting 62 rules. 

The \gls{CRS} rules are sub-divided into four sets, each identified by a specific \emph{\gls{PL}}. These sets are constructed such that PL1 $\subset \ldots \subset$ PL4, \ie, increasing the PL amounts to include more \gls{CRS} rules, with PL4 including all of them. In practice, higher PLs tend to exhibit a higher detection rate but also cause a higher number of false alarms.
Within each PL, rules are assigned a specific weight, referred to as their \emph{severity level}.
The severity level of each rule is assigned by domain experts, based on their subjective evaluation of the potential impact of the attack that such a rule aims to prevent.
Then, the score associated with each incoming request is computed as the sum of the severity levels associated with the firing rules. If such a score exceeds a given threshold, the incoming request is blocked. %In the following, we will refer to the PL1-PL4 configurations of the \gls{CRS} as different \modsecurity configurations. 
More details on how the PL1-PL4 CRS configurations work are provided in \autoref{sec:background}.

While the domain knowledge poured in developing the \gls{CRS} rules is extremely valuable, in this work we
use the well-known \modsecurity WAF~\cite{modsecurity_book} equipped with the CRS rules to show that the heuristic choices made to select and combine such rules can lead to: (i) a suboptimal trade-off between detection rate and false alarms; and (ii) a substantial lack of robustness to \emph{adversarial} \gls{SQLi} attacks, \ie, functionality-preserving manipulations of the \gls{SQLi} attack payload aimed to evade detection~\cite{demetrio2020waf,scano2024modsec}.
To overcome these limitations, we then propose a novel robust machine learning approach to selecting and combining the CRS rules, named \advms, which is conceptually represented in \autoref{fig:system_overview} and detailed in \autoref{sec:robustml}. This approach is built upon two main contributions.
First, we propose using \gls{ML} to automate both rule selection and weighting, adjusting the \gls{CRS} configuration based on the traffic data collected from the monitored web services, and building on our preliminary findings in~\cite{scano2024modsec}. The underlying idea is to train a linear \gls{ML} model using \textit{all} the \gls{CRS} rules as input features, aiming to improve the trade-off between detection and false alarm rates, by specializing the model to the specifics of the traffic data collected from the monitored applications. Furthermore, enforcing a sparse regularization during training enables the selection of an effective subset of rules as a by-product, avoiding the need for manual selection of the \gls{CRS} rules in each PL.
The second main contribution of this work is the definition of a novel adversarial training scheme to improve model robustness against adversarial \gls{SQLi} attacks. 
To craft these attacks, we leverage \wafamole~\cite{demetrio2020waf}, \ie, a black-box mutational fuzzer~\cite{fuzzingbook2023:MutationFuzzer} that iteratively selects the best combination of random manipulations of SQLi payloads to reduce their probability of being detected by the targeted \gls{WAF}.
We also show in \autoref{sec:advmodsec} that using $\ell_\infty$-norm regularization on a linear model yields equivalent robust solutions, avoiding the computational burden of optimizing attacks during training.

Through our experiments, reported in \autoref{sec:experiments} and conducted on two publicly-available datasets~\cite{demetrio2020waf,scano2024modsec}, we show that \advms overcomes the limitations of the current \gls{CRS}, by detecting 30\% attacks more with much fewer rules.
\advms also provides an unprecedented level of robustness against adversarial \gls{SQLi} attacks, i.e., 85\% more than the default \gls{CRS} configurations.
By deepening the investigation of this result, we discover that \advms gives more emphasis to rules that are (i) less affected by adversarial \gls{SQLi} attacks and (ii) accidentally triggered by side-effect artifacts introduced by the adversarial \gls{SQLi} manipulations. 

To conclude, we remark that our work is the first to demonstrate the effectiveness of adversarial training in the \glspl{WAF} domain (specifically, for detecting \gls{SQLi} attacks) when leveraging state-of-the-art input-space \gls{SQLi} manipulations.
This is completely different from other domains like image classifiers, where adversarial training is not sufficient to achieve a high level of robustness, and the effective mitigation of the risks presented by adversarial examples is still an open issue. 
For this reason, we firmly believe that our work provides interesting and novel insights on how to design robust machine learning models for cybersecurity.
We discuss these aspects along with related work in \autoref{sec:related_work}, and the limitations of our approach as well as the corresponding future research directions in \autoref{sec:conclusions}. 
We have also publicly released our code to foster reproducibility of our work.\footnote{\url{https://github.com/pralab/modsec-advlearn}}
\section{Background}\label{sec:background}
We introduce here \gls{SQLi} attacks and the \gls{OWASP} \gls{CRS} project. We then describe how to generate adversarial \gls{SQLi} attacks using \sota fuzzing techniques.

\subsection{SQL Injection (SQLi)}\label{sec:sqli}
These attempts to retrieve or alter sensitive information from a target database, modifying data without authorization, or even execute privileged operations on the database~\cite{appelt2017tuningwaf}.
This can be achieved via specific SQL code fragments that are passed in the original request.
If the application does not sanitize the user input and simply concatenates it with the query, the SQL fragment is interpreted as part of the original SQL query.
The login form of a web application is a paradigmatic example (\autoref{code:sql_query}). 
The credentials of a user are provided in two input fields (e.g., \texttt{\$user} and \texttt{\$passwd}) and sent via an HTTP request.
The credentials are then checked server-side via a database query.
However, a malicious user injects SQL fragments in the \texttt{\$user} parameter, \emph{e.g.}, "\texttt{admin'-- }".
As shown in \autoref{code:sqli_example}, this bypasses the original SQL query's password check, resulting in a successful \gls{SQLi} attack, allowing login with just a valid username.
%As shown in \autoref{code:sqli_example}, this comments out the portion of the original SQL query that checks for a valid password, leading to a successful \gls{SQLi} attack, \ie, the login is performed using only a valid username, without requiring the password.

\begin{lstlisting}[
   language=SQL,
   basicstyle=\ttfamily,
   commentstyle=\color{gray},
   showspaces=false,
   showstringspaces=false,
   breakindent=1em,
   breaklines=true,
   % xleftmargin=1.8em,
   keepspaces=true,
   frame=single,
   caption={Example of SQL query vulnerable to injection.},
   captionpos=b,
   label={code:sql_query}
]
SELECT * FROM users WHERE username = '$user' AND password = '$passwd'
\end{lstlisting}

\begin{lstlisting}[
   language=SQL,
   basicstyle=\ttfamily,
   commentstyle=\color{gray},
   showspaces=false,
   showstringspaces=false,
   breakindent=1em,
   breaklines=true,
   % xleftmargin=1.8em,
   keepspaces=true,
   frame=single,
   caption={Example of SQL Injection on \autoref{code:sql_query}.},
   captionpos=b,
   label={code:sqli_example}
]
SELECT * FROM users WHERE username = 'admin'-- ' AND password = 'x'
\end{lstlisting}

\subsection{The OWASP Core-Rule-Set (CRS) Project}\label{sec:owasp_crs}
This open-source initiative is one of the most widely-used sets of detection rules targeting OWASP Top 10 web security risks \cite{owasp_top_ten}.
It is not only the reference rule set of several open-source \gls{WAF} solutions like \modsecurity~\cite{modsecurity_book} and Coraza,\footnote{\url{https://coraza.io}} but is also adopted in many commercial solutions including Google Cloud Armor, Microsoft Azure, and Cloudflare \glspl{WAF}~\cite{crs_doc}.

\myparagraph{Detection Rules.}
These are regular expressions (regex) that match specific byte patterns in requests.
For instance, line 3 of \autoref{code:sqli_rule} captures several patterns of comments commonly used in \gls{SQLi} attacks such as "\textbf{\texttt{;--}}" and "\textbf{\texttt{-- }}".
Each rule is denoted by a unique identifier (\verb|id| in line 4), whose suffix also indicates the type of attack it is designed to identify (those starting with 942 target \gls{SQLi} attacks~\cite{crs_doc}).
Notable among configuration settings are the Paranoia Level (line 8) and Severity Level (line 9), which are explained below.

\myparagraph{Paranoia Level.}
It defines the set of rules that are enabled to analyze the incoming HTTP requests~\cite{crs_doc}.
The \gls{CRS} includes four \glspl{PL} (PL1 - PL4) and each rule is assigned to a specific \gls{PL}; e.g., the rule in~\autoref{code:sqli_rule} belongs to PL1 (line 8).
Moreover, rules are grouped together by \gls{PL} in a nested way: setting a certain \gls{PL} enables all the rules assigned to that \gls{PL}, as well as those assigned to lower \glspl{PL}.
For instance, PL3 enables all the rules related to such \gls{PL}, as well as those assigned to PL1 and PL2. Consequently, PL4 will enable all the rules.

\myparagraph{Severity Level.}
Each CRS rule is heuristically given a \emph{severity level}, i.e., a positive integer that quantifies how severe the corresponding attack could be~\cite{crs_doc}.
The \gls{WAF} applies the rules on each incoming request, and sums the severity levels of the firing rules.
If the aggregated score exceeds a predefined threshold, the request is flagged as malicious.
The \gls{CRS} defines four severity levels: CRITICAL (5), ERROR (4), WARNING (3) and NOTICE (2); e.g., the severity level of the rule in~\autoref{code:sqli_rule} is CRITICAL (line 9), so it contributes to the aggregated score with a value of 5.

\begin{lstlisting}[
   numbers          = left,
   stepnumber       = 1,
   xleftmargin      = 3.0em,
   showspaces       = false,
   basicstyle       = \ttfamily\scriptsize,
   breaklines       = true,
   showstringspaces = false,
   keepspaces       = true,
   frame            = single,
   caption          = {CRS rule  detecting typical comments in \gls{SQLi}.},
   captionpos       = b,
   label            = {code:sqli_rule}
]
SecRule REQUEST_COOKIES|!REQUEST_COOKIES:/__utm/
|REQUEST_COOKIES_NAMES|ARGS_NAMES|ARGS|XML:/* 
"@rx (?i)/\*[\s\v]*?[!\+](?:[\s\v\(-\)\-0-9=A-Z_a-z]+)?\*/" \
  "id:942500,
  block,
  msg:'MySQL in-line comment detected',
  tag:'attack-sqli',
  tag:'paranoia-level/1',
  severity:'CRITICAL',
  setvar:'tx.sql_injection_score=+
    %{tx.critical_anomaly_score}',
  setvar:'tx.inbound_anomaly_score_pl1=+
    %{tx.critical_anomaly_score}'"
\end{lstlisting}

\subsection{Adversarial SQLi Attacks against WAFs}
\label{sec:advsqli}
In the context of \glspl{WAF}, the problem of finding \gls{SQLi} attacks that can bypass the target \gls{WAF} is \emph{adversarial in nature}.
To this end, the attacker may manipulate \gls{SQLi} attack payload to evade detection while preserving its malicious functionality~\cite{demetrio2020waf,hemmati2022rlwaf}.
For instance, the \gls{SQLi} rule reported in~\autoref{code:sqli_rule} can detect the following \gls{SQLi} payload: \textbf{\texttt{admin'}} \texttt{OR 1=1; --'}.
However, by inserting a white space character (\texttt{' '}) in the original attack payload, we generate a semantically-equivalent \gls{SQLi} attack: \textbf{\texttt{admin'}} \texttt{OR 1=1; --'}, that can evade the rule.

\myparagraph{WAF-a-MoLE.}
\begin{table}[t]
 \caption{Manipulation functions applied by \wafamole.}
    \resizebox{\linewidth}{!}{%
        \begin{tabular}{l l}
            \toprule
            Manipulation & Effect on payload \\
            \midrule
            Case Swapping &  CS(\texttt{admin' OR 1=1}) $ \rightarrow $ \texttt{ADmIn' oR 1=1} \\
            Whitespace Substitution & WS(\texttt{admin' OR 1=1}) $\rightarrow$ \texttt{admin'\textbackslash n OR \t 1=1}\\
            Comment Injection & CI(\texttt{admin' OR 1=1}) $\rightarrow$ \texttt{admin'/**/OR 1=1} \\
            Comment Rewriting & CR(\texttt{admin'/**/OR 1=1}) $\rightarrow$ \texttt{admin'/*abc*/OR 1=1} \\
            Integer Encoding & IE(\texttt{admin' OR 1=1}) $\rightarrow $ \texttt{admin' OR 0x1=1} \\
            Operator Swapping & OS(\texttt{admin' OR 1=1}) $\rightarrow $ \texttt{admin' OR 1 LIKE 1} \\
            Logical Invariant & LI(\texttt{admin' OR 1=1}) $\rightarrow $ \texttt{admin' OR 1=1 AND 2<>3} \\
            \bottomrule
        \end{tabular}
    }
    \label{tab:h_wafamole}
\end{table}
Our methodology builds upon \wafamole~\cite{demetrio2020waf}, a state-of-the-art, open-source \gls*{SQLi} guided mutational fuzzer~\cite{fuzzingbook2023:MutationFuzzer}, aimed at finding semantically-equivalent SQLi attacks that evade detection through the application of functionality-preserving manipulations. 
%that, by querying the target \gls{WAF}, allows to generate optimized adversarial \gls{SQLi} by applying a set of semantic-preserving mutation operators to a given \gls{SQLi} payload.
These manipulations, detailed in \autoref{tab:h_wafamole}, can be encoded as a function $h(\vct z, \vct \delta)$, where $\vct z$ is the SQLi query to be modified, and $\vct \delta$ are the parameters defining the perturbation.
For instance, \wafamole can include new comments into the SQLi, adding always-true or always-false statements, converting numbers to a different base, or replacing them with SQL commands that, once evaluated, produce the same number.
Such choice is controlled by $\vct \delta$, which specifies the type of manipulation and the content that should be injected or replaced.
\wafamole then iteratively refines the choice of $\vct \delta$ to decrease the confidence of the targeted WAF into classifying the modified SQLi payload as an attack.
This is achieved by generating several candidates through random choices of $\vct \delta$ in each iteration, and retaining only those that successfully reduce the confidence score attributed by the WAF.
In this work, we will use \wafamole (i) to show that the standard configurations of the \gls{CRS} can be bypassed by optimizing adversarial \gls{SQLi} attacks against them, and (ii) to generate adversarial SQLi queries for our novel \textit{problem-space} adversarial training approach.\footnote{We refer to our approach as problem-space adversarial training to differentiate it from approaches that simulate the effect of attacks by modifying only their feature vectors, without even producing the actual samples~\cite{pierazzi2020intriguing}.}
\section{Robust Machine Learning against Adversarial SQLi Attacks}
\label{sec:robustml}

We detail here how we design our robust \gls{ML} approach to (i) improving the tradeoff between detection and false alarm rates by optimizing the selection and combination of the \gls{CRS} rules (\autoref{sec:mlmodsec}), and (ii) improving robustness to adversarial SQLi attacks (\autoref{sec:advmodsec}).

\subsection{\mlms: Machine Learning for \gls{CRS}}
\label{sec:mlmodsec}
We start by discussing the building block developed from our initial findings, i.e., \mlms (step 1 of~\autoref{fig:system_overview}).
It consists of two components: (i) a feature extraction phase that encodes the \gls{CRS} rules into a vector representation; and (ii) a \gls{ML} model that learns how to optimally combine the \gls{CRS} rules, avoiding the manual tuning of their severity scores.

\myparagraph{Detection Rules as Features.}
The input space is represented by SQL queries that are classified as malicious or benign by a \gls{ML} model.
Each SQL query is a string of readable characters, represented as $\vct z \in \set Z$, being $\set Z$ the space of all possible queries.
Let $\set D$ be the set of selected \gls{SQLi} rules from \gls{CRS}, and $\con d = |\set D|$ its cardinality.
We denote with $\phi : \set Z \mapsto \set X = \{0,1\}^{\con d}$ a function that maps a SQL query $\vct z$ to a $\con d$-dimensional Boolean feature vector $\vct x = ( \phi_1(\vct z), \ldots, \phi_{\con d}(\vct z))$, %\in \set X = \{0,1\}^{\con d},
where each $\phi_j(\vct z)$ corresponds to evaluating the \textit{j}-th \gls{SQLi} rule on the input query $\vct z$.
Each $\phi_j(\vct z)$ returns 1 if the corresponding rule has been triggered by the SQL query $\vct z$, and 0 otherwise.

\myparagraph{Optimal Combination of \gls{CRS} Rules with ML.}
\label{sec:ml_models}
To optimally tune the contribution of the \gls{CRS} rules towards effectively classifying the input requests we leverage three different \gls{ML} algorithms on the feature representation defined above: two linear models, i.e., \glspl{SVM}~\cite{vapnik95} and \gls{LR}~\cite{logistic}, both with $\ell_1$ and $\ell_2$ regularization; and a non-linear \gls{RF} model~\cite{breiman2001rf}.

Linear models are particularly valuable as they can automatically adjust the severity score assigned to each rule, rather than relying on default \gls{CRS} values, whereas non-linear models may give us an indication of the best performance achievable. Furthermore, when sparse ($\ell_1$) regularization is applied, linear models can effectively select an optimal subset of \gls{CRS} rules, potentially eliminating the need for predefined \glspl{PL}. Although our approach is applicable to both linear and non-linear models, integrating a non-linear model within the existing \gls{CRS} rules may pose additional complexity and scalability issues, while also worsening the interpretability of the \gls{WAF} decisions. Conversely, the weights learned by any linear model can be directly plugged-in into the \gls{CRS} rules without any significant disadvantage. Let us finally remark that, compared to our previous work in~\cite{scano2024modsec}, we extend the evaluation of \mlms models by assessing their robustness against adversarial attacks, and also considering an additional dataset~\cite{demetrio2020waf}.

\subsection{\advms: Robustness against Adversarial SQLi}\label{sec:advmodsec}
While \modseclearn can learn to automatically select and combine the CRS rules from the training data, yielding a better tradeoff between detection and false alarm rates, it may not guarantee a sufficient degree of robustness against adversarial \gls{SQLi} attacks.
Hence, we detail here how we extend \mlms to improve robustness against adversarial SQLi attacks. We refer to this approach as \advms (step 2 of~\autoref{fig:system_overview}).
The underlying idea is to leverage \emph{adversarial training} (AT)~\cite{madry2018advtrain,biggio2018wild} to include adversarial SQLi examples during training, thereby giving the model the ability to withstand the corresponding evasive attack patterns at test time. 

In the common setting of image classification, AT leverages gradient-based attacks to craft adversarial examples, as the corresponding optimization problem is end-to-end differentiable, and  the considered perturbations are simply additive. 
This is not directly applicable in the case of SQLi attacks, as well as in other domains~\cite{Demontis2017YesML,demetrio2021tifs}, in which models are not end-to-end differentiable (given the presence of non-differentiable pre-processing and feature extraction steps) and perturbations are not additive.
We thus consider here two distinct approaches to performing AT: \textit{feature-space} and \textit{problem-space} AT.

\myparagraph{Feature-space AT.} In this setting, we make the na\"ive assumption that each adversarial SQLi attack can enable or disable up to a number $\lambda$ of CRS rules to evade the target WAF.
This amounts to optimizing the following min-max objective:
\begin{equation}
    \min_{\vct{w}} \max_{\|\vct{\delta}_i\|_1 \le \lambda} \sum_{i} \mathcal{L}(y_{i},f_{\vct{w}}(\vct x_i + \vct{\delta}_i)) \, ,
\label{eq:rob_problem}
\end{equation} 
where (i) $\vct x_i = \phi(\vct z_i)$ is the $\con d$-dimensional Boolean vector representing the activations of the CRS rules for the given SQL sample $\vct z_i$; 
(ii)~$y_i \in \set Y = \{-1,+1\}$ is its label;
(iii)~$f_{\vct w} : \set X \rightarrow \mathbb R$ is the ML model, parameterized by $\vct w$, which classifies a sample as positive if $f_{\vct w}(\vct x) \ge 0$, and as negative otherwise;
(iv) $\mathcal{L} : \set Y \times \mathbb R \rightarrow \mathbb{R}$ is the loss function to be minimized;  and 
(v) $\vct \delta_i$ is the manipulation that switches on or off at maximum $\lambda$ rules from the CRS (expressed as an $\ell_1$ norm constraint).
Furthermore, it must also hold that $\vct x_i + \vct \delta_i \in \set X = \{0,1\}^{\con d}$, as the activations have to remain Boolean also after perturbation. 
In practice, the min-max problem is solved iteratively. In each iteration, the inner problem amounts to finding adversarial examples against the given model, while the outer problem adjusts the model parameters $\vct w$ to re-classify them correctly.\footnote{Note that, for simplicity, in the given formulation we consider adversarial modifications of both benign and malicious training samples, but this can be easily adjusted to consider only manipulations of malicious SQLi queries.}

In this work, we do not solve the problem given in \autoref{eq:rob_problem} directly, as it is too computationally demanding. Instead, we derive an equivalent formulation for linear \glspl{SVM} based on solving the inner problem in closed form, which simply amounts to using a different regularization term.

\textit{Robustness through Regularization with SecSVM.} As originally shown by by Xu~\etal~\cite{xu2009robustness}, and subsequently adapted in~\cite{Demontis2017YesML} to the case of Android malware detection, the inner problem in \autoref{eq:rob_problem} can be solved in closed form when the loss function $\mathcal{L}$ is \textit{linear}. This is the case, e.g., when using the hinge loss and a linear model $f_{\vct w}(\vct x) = \vct w^t \vct x + b$, as in linear \gls{SVM} training.
In this case, the inner problem in \autoref{eq:rob_problem} for each sample can be rewritten as:
\begin{equation}
    \max_{\| \vct \delta \|_1 \leq \lambda} \mathcal L(y, f_{\vct{w}}(\vct{x})) + \vct \delta^\T \nabla \mathcal L(y, f_{\vct{w}} \,(\vct{x})) \, ,
\end{equation}
where the gradient $\nabla \mathcal L(y, f_{\vct{w}}(\vct x))$ corresponds to the weight vector $\vct w$. The above problem then amounts to maximizing a scalar product over an $\ell_1$-norm constraint, i.e., $\max_{\| \vct \delta \|_1 \leq \lambda} \vct \delta^\T \vct w$, whose solution is proportional to the dual norm of the weight vector, given as $\lambda \|\vct w\|_\infty$~\cite{xu2009robustness}. This means that we can rewrite the robust (min-max) optimization problem given in \autoref{eq:rob_problem} as a much simpler regularized problem:
\begin{equation}
    \min_{\vct{w}}\sum_{i} \mathcal{L}(y_i, f_{\vct{w}}(\vct{x}_i)) + \lambda \|\vct{w}\|_{\infty} \, .
\label{eq:reg_problem}
\end{equation} 
This finding sheds light on the role of the regularization term, showing that its choice should be based on the type of noise that affects the input data.
In particular, it tells us that $\ell_\infty$ is the optimal regularizer for training a robust model against $\ell_1$-norm (sparse) perturbations. It is also not difficult to see that, analogously, the standard $\ell_2$-norm~SVM is optimal against $\ell_2$-norm (dense) perturbations~\cite{xu2009robustness,Demontis2017YesML}. 

The above problem can be equivalently re-parameterized by (i) replacing the regularization parameter $\lambda$ with the hyperparameter $t$, i.e., bounding the weight values $\vct w$ in $[-t, t]$; and (ii) introducing the slack variables $\vct{\xi}$ to measure how far each sample $\vct{z}_i$ is from being correctly classified:
\begin{eqnarray}
    \label{eq:lin_problem} \min_{\vct{w}, \xi} && \sum_{i} \xi_i \\
    \mathrm{s.t.} && \xi_i \ge 1-y_i f_{\vct{w}}(\vct{x}_i),\quad \forall i \in 1,\dots,n \\
    && \xi_i \ge 0,\quad \forall i \in 1,\dots,n \\
    && -t \le w_j \le t,\quad \forall j \in 1,\dots,d \, .
\end{eqnarray}
The given formulation corresponds to a linear programming problem in its canonical form, which can be solved using standard, off-the-shelf solvers, such as the simplex algorithm or interior-point methods. In this work, the optimization problem is solved using the linear programming solver provided by the SciPy library\footnote{\url{https://docs.scipy.org/doc/scipy/reference/generated/scipy.optimize.linprog.html}}. We refer to this robust learning approach as \gls{SecSVM} in our experiments.\footnote{Note that, even if we keep the same name of the approach proposed in~\cite{Demontis2017YesML}, our SecSVM implementation is different, as we are neither using custom bounds on each feature value nor any $\ell_2$ regularization.}

\myparagraph{Problem-space AT with \wafamole.} Let us now define a different approach to learning robust models directly against problem-space perturbations.
The reason is that hardening models via feature-space AT or with ad-hoc regularization methods may provide an overly pessimistic approach to feature manipulation that does not consider the specific constraint of realizable, semantic- and functionality-preserving \gls{SQLi} attacks.
In particular, practical \gls{SQLi} manipulations may not enable switching on or off individual CRS rules independently, and may inadvertently trigger certain rules even if the attack aims to bypass the detection. 
Furthermore, such manipulations cannot be modeled as additive perturbations, and computing them typically requires inverting a complicated, non-differentiable feature extraction step; in our case, this would amount to reversing the inner workings of each CRS rule. For this reason, feasible SQLi attacks, as many other adversarial perturbations aimed to bypass ML models for cybersecurity-related tasks~\cite{demetrio2021tifs}, cannot be typically optimized via gradient descent directly. To overcome these issues, we consider here a more general problem-space AT procedure, defined as:
\begin{equation}
    \min_{\vct{w}} \max_{\vct{\delta}_i \in \Delta} \sum_{i} \mathcal{L}(y_{i},f_{\vct{w}}(\phi(h(\vct z_i, \vct \delta_i))) \, ,
\label{eq:rob_problem2}
\end{equation} 
where $h(\vct z, \vct \delta)$ is a manipulation function that modifies the input SQL query $\vct z$ and returns a semantic-preserving SQL query $\vct z^\prime$, based on the choice of its input parameters $\vct \delta \in \Delta$.
The set $\Delta$ constrains the input manipulations described in \autoref{tab:h_wafamole} to produce valid samples, e.g., picking only visible characters when adding or re-writing comments, produce always-true or always-false conditions that do not change the original evaluation of the payload, or encoding integers in a specific base different from the original one.

To solve the problem given in \autoref{eq:rob_problem2}, we consider a gradient-free \emph{black-box} optimization achieved through \wafamole as shown in~\autoref{algo:adv_train}, modifying only the set of malicious SQLi payloads, and leaving benign SQL queries unchanged.
\begin{algorithm}[t]
    \SetKwInOut{Input}{Input}
    \SetKwInOut{Output}{Output}
    \Input{$\set D = (\vct z_i, y_i)_{i=1}^M$, the training set of SQL samples with labels;
           $f$, the ML model;
           $\mathcal{L}$, the loss function;
           $N$, the number of adversarial SQLi attacks to be added to the initial training set.
           }
    \Output{$f_{\vct w^\star}$, the model with re-trained parameters $\vct w^\star$}
    $\set Z^{\prime} \leftarrow \{\vct z_i\}_{i=1}^N \text{ with } \vct z_i \sim \set D \text{ s.t. } y_i = +1$\label{line:adv_train_1} \\
    \textbf{for} $ \vct z $ \textbf{in} $ \set Z^{\prime}$\\
    \Indp
       $\vct z^\star \leftarrow $ \emph{\wafamole}$(\vct z, f)$\label{line:adv_train_2} \\
      $\set Z \leftarrow \set Z \cup \{ \vct z^\star \}$; $\set Y \leftarrow \set Y \cup \{ +1 \};$\label{line:adv_train_3} \\
    \Indm
    $\vct w^\star \leftarrow \argmin_{\vct w} \frac{1}{|\set Z|}\sum_{i=0}^{|\set Z|}\mathcal{L}(y_i, f_w(\vct z_i))$\label{line:adv_train_4} \\
    return $f_{\vct w^\star}$
    \caption{Adversarial training of \advms with \wafamole}
    \label{algo:adv_train}
\end{algorithm}
Given a dataset $\set Z$ of benign queries and \gls{SQLi} attacks labeled as 0 and 1 respectively, we first create a new set ($\set Z^{\prime}$) by randomly sampling a given amount of \gls{SQLi} from the training dataset (\autoref{line:adv_train_1}).
Then, for each \gls{SQLi} sample of this newly created set, we use \wafamole (\autoref{sec:advsqli}) to generate the corresponding adversarial \gls{SQLi} (\autoref{line:adv_train_2}) and add it to the training data with its malicious (+1) label (\autoref{line:adv_train_3}).
The parameters of the model are finally optimized on the training set including the adversarial \gls{SQLi} samples (\autoref{line:adv_train_4}).

In our experiments, we will also retrain \gls{SecSVM} with the proposed problem-space AT method, to consider less pessimistic, but practical, adversarial attacks. In principle, this should allow us to further improve the robustness-detection tradeoff of \gls{SecSVM} against real-world \gls{SQLi} attacks. We would like to finally remark that, even if we leverage an existing tool like \wafamole to this end, to our knowledge, this is the first attempt to define a novel problem-space adversarial training approach to hardening ML-based \glspl{WAF} against practical adversarial SQLi attacks.

\section{Experimental Analysis}
\label{sec:experiments}

We report here three different experiments to validate our methodology.
First, we evaluate the detection capabilities of the \gls{CRS}.
This is achieved by using the vanilla \modsecurity \gls{WAF} as the underlying engine (\autoref{sec:exp-modsecurity}), showing that its na\"ive approach of combining the \gls{CRS} rules based on manually-assigned weights is largely suboptimal and significantly vulnerable to adversarial \gls{SQLi} attacks.
Second, we empirically show that the ML-based tuning adopted by \mlms allows one to fill the gaps of the vanilla \modsecurity, by significantly enhancing its detection rate up to 30\%, and we continue highlighting how \mlms enhances the performances thanks to the adaptation of weights, while also reducing the number of rules needed (\autoref{sec:exp-ml}).
Third, we present the results of our novel adversarial training approach showing that \advms is 85\% more robust than \modsecurity (\autoref{sec:exp-adv}).

\subsection{Experimental Setup}\label{sec:exp_setup}
In this section, we describe the two datasets used in our analysis, along with the setup of the \modsecurity WAF, the \wafamole fuzzer, and the ML models used.

\myparagraph{Datasets}.
We conduct our experiments using two datasets. The first one, \wafamole Dataset~\cite{demetrio2020waf}, which consists of 393,629 malicious and 345,199 benign SQL queries. Benign samples were generated from a restricted SQL grammar, while attacks were generated using well-known web security testing tools such as SQLmap and OWASP ZAP~\cite{demetrio2020waf}. The second one, \modseclearn Dataset~\cite{scano2024modsec}, instead consists of 25,000 malicious SQLi payloads and 25,000 benign HTTP requests, based on real-world traffic. Legitimate samples were collected from the \textit{open-appsec} dataset,\footnote{\url{https://github.com/openappsec/waf-comparison-project/tree/main/Data}}, which contains samples from various real-world scenarios. Malicious samples were collected from multiple sources, and generated through security testing tools such as SQLmap, by executing it with different tampering scripts designed for payload obfuscation.

We divide each dataset into four subsets: training (\texttt{train}), test (\texttt{test}), adversarial training (\texttt{train-adv}), and adversarial test (\texttt{test-adv}). \autoref{tab:dataset} shows the distribution of samples across the four subsets for each dataset: \texttt{train} contains an equal number of benign and SQLi queries, and it is used to train our target \glspl{WAF}; \texttt{test}, disjoint from \texttt{train}, is used to evaluate the baseline performances of target \glspl{WAF}, including vanilla \modsecurity, \mlms, and \advms, across different \glspl{PL}; \texttt{train-adv} contains the same samples of \texttt{train}, but  $50\%$ of the \gls{SQLi} queries in the \wafamole dataset and the $25\%$ of the \gls{SQLi} queries in the \modseclearn dataset are optimized using \wafamole against the target \gls{WAF} for problem-space AT; \texttt{test-adv} contains the same samples of \texttt{test}, but optimizes all the \gls{SQLi} queries using \wafamole to bypass the target \gls{WAF}, i.e., to evaluate its robustness. 
We would like to clarify that, although using the same \gls{SQLi} fuzzer (\ie, \wafamole), the adversarial examples generated for building the adversarial training and test sets are different. They are independently optimized against each target model at test time, resulting in the application of different, optimal manipulation strategies. Thus, the sets are independent, ensuring an unbiased evaluation.

%\begin{table}[!htb]
%\caption{Number of legitimate, SQLi, and adversarial SQLi samples in each \texttt{train}, \texttt{test}, \texttt{train-adv}, and \texttt{test-adv} subset, for the WAF-a-MoLE and ModSec-Learn datasets.}
%\resizebox{\linewidth}{!}{%
%    \begin{tabular}{c c c c c c}
%    \toprule & \multicolumn{2}{c}{\texttt{train}} & \multicolumn{3}{c}{\texttt{train-adv}} \\
%    \cmidrule(lr){2-3} \cmidrule(lr){4-6}
%    & Legitimate & \gls{SQLi} & Legitimate & \gls{SQLi} & Adversarial \gls{SQLi} \\
%    \midrule
%    WAF-a-MoLE & $10,000$ & $10,000$ & $10,000$ & $5,000$ &  $5,000$ \\
%    ModSec-Learn & $20,000$ & $20,000$ & $20,000$ & $15,000$ & $5,000$ \\
%    \bottomrule
%    \end{tabular}
%}
%\label{tab:dataset}
%\end{table}

\begin{table}[!htb]
\caption{Number of legitimate, SQLi, and adversarial SQLi samples in \texttt{train}, \texttt{test}, \texttt{train-adv}, and \texttt{test-adv}, for the WAF-a-MoLE and ModSec-Learn datasets.}
\resizebox{\linewidth}{!}{%
    \begin{tabular}{c c c c c c}
    \toprule 
    & \multicolumn{2}{c}{\texttt{train}} & \multicolumn{3}{c}{\texttt{train-adv}} \\
    \cmidrule(lr){2-3} \cmidrule(lr){4-6}
    & Legitimate & \gls{SQLi} & Legitimate & \gls{SQLi} & Adversarial \gls{SQLi} \\
    \midrule
    WAF-a-MoLE   & $10,000$ & $10,000$ & $10,000$ & $5,000$ & $5,000$ \\
    ModSec-Learn & $20,000$ & $20,000$ & $20,000$ & $15,000$ & $5,000$ \\
    \midrule
    & \multicolumn{2}{c}{\texttt{test}} & \multicolumn{3}{c}{\texttt{test-adv}} \\
    \cmidrule(lr){2-3} \cmidrule(lr){4-6}
    & Legitimate & \gls{SQLi} & Legitimate & \gls{SQLi} & Adversarial \gls{SQLi} \\
    \midrule
    WAF-a-MoLE   & $2,000$ & $2,000$ & $2,000$ & - & $2,000$ \\
    ModSec-Learn & $5,000$ & $2,000$ & $5,000$ & - & $2,000$ \\
    \bottomrule
    \end{tabular}
}
\label{tab:dataset}
\end{table}

\myparagraph{\modsecurity and \wafamole Setup.}
We evaluate \modsecurity v3.0.10 with the \gls{CRS} v4.0.0. 
Since we focus on the detection of \gls{SQLi} attacks, we only enable its \gls{SQLi} rules.\footnote{\url{https://github.com/coreruleset/coreruleset/blob/v4.0.0/rules/REQUEST-942-APPLICATION-ATTACK-SQLI.conf}}
We configure \wafamole to use a maximum of $2,000$ queries, to ensure convergence of the attack optimization. Attacks are optimized by minimizing the confidence score assigned to the malicious class by the targeted WAF. 

\myparagraph{Implementation Details.} We use \textit{pymodsecurity} v0.1.0,\footnote{\url{https://github.com/pymodsecurity/pymodsecurity}} as our Python interface to \modsecurity. To efficiently query and test \modsecurity, we have extended \wafamole by developing a dedicated \textit{pymodsecurity} interface, which avoids instantiating the whole web server. This interface is available in our open-source repository. 

\myparagraph{Machine Learning.}
We leverage scikit-learn v1.4.0 implementations of \gls{SVM} (LinearSVC), \gls{LR}, and \gls{RF} to implement both \modseclearn and \advms.
For the \gls{SVM} and \gls{LR} models, we experiment with both $\ell_1$ and $\ell_2$ regularizers. 
We implement \gls{SecSVM} as described in \autoref{sec:advmodsec}, using the linear programming solver provided by SciPy. We refer to it as \emph{\mlms SecSVM} in the reported tables. The hyperparameters of each model are tuned via grid search, performing a 5-fold cross validation on the training set (\texttt{train}) to maximize the mean F1 score.  In the case of \modseclearn, for the \gls{SVM} and \gls{LR}, we tune the regularization parameter $C \in \{ 10^{-3}, 10^{-2}, 10^{-1}, 0.5, 1.0 \}$. The best value found is typically $C = 0.5$ for both models. 
For \gls{SecSVM}, we tune the hyperparameter $t \in \{0.1, 0.2, \ldots, 1.0\}$. The best $t$ value is typically found to be $0.5$.
The \gls{RF} model is used with its default hyperparameters.
In the case of \advms, adversarial training is applied only for PL4, given that the models trained on this \gls{PL} demonstrated better performance on the test set (see~\autoref{sec:exp-ml}). The hyperparameters are tuned using the same procedure described in this paragraph, finding approximately the same best values, except for \gls{SecSVM}, for which $t=1.0$ yields better results.

\begin{figure*}[htbp]
\centering
\includegraphics[width=\textwidth]{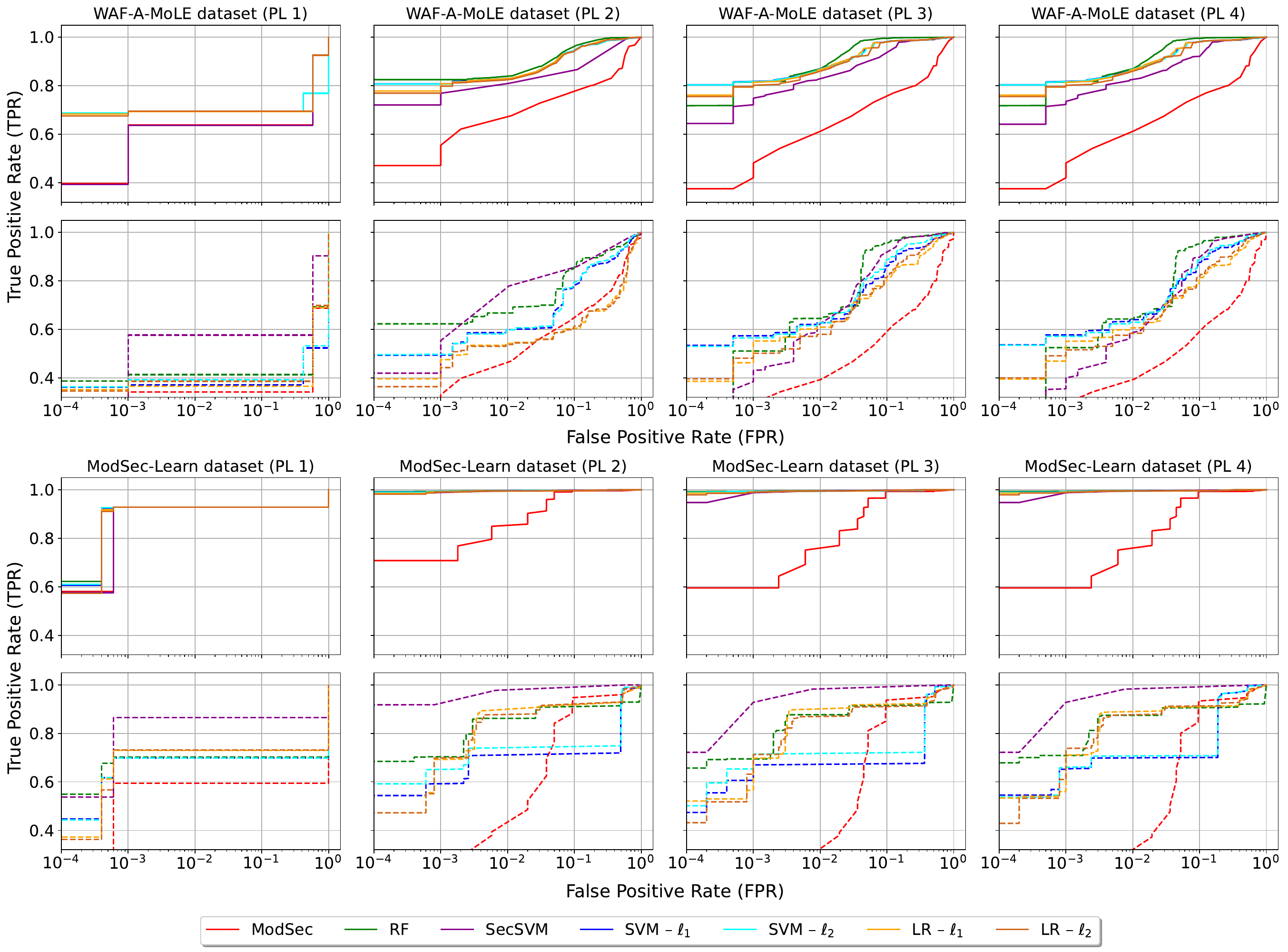}
\caption{ROC curves of vanilla \modsecurity (ModSec) and \mlms approaches (\gls{SVM}, \gls{RF}, \gls{LR}, and \gls{SecSVM}), evaluated on \texttt{test} (solid lines) and \texttt{test-adv} (dashed lines), for \wafamole (\textit{first two rows}) and \mlms (\textit{last two rows}) datasets. Each curve reports the fraction of detected \gls{SQLi} attacks against the fraction of misclassified legitimate requests.}
\label{fig:roc}
\end{figure*}

\subsection{Evaluation of \modsecurity}
\label{sec:exp-modsecurity} 
The first goal of our experimental analysis is to understand the detection capability of the vanilla \modsecurity. Rather than focusing only on \gls{CRS} default values, we experiment with it over its entire configuration space, considering all the possible values for the \glspl{PL} and the classification threshold. 
Hence, for each \gls{PL}, we compute the Receiver-Operating-Characteristic (ROC) curve, which reports the detection rate, a.k.a. True Positive Rate (TPR, \ie, the fraction of correctly-detected malicious \gls{SQLi} requests) against the False Positive Rate (FPR, \ie, the fraction of wrongly-classified legitimate requests) obtained by considering all possible classification threshold values. We report our findings with red lines in~\autoref{fig:roc}, while in~\autoref{tab:tpr_table} we extrapolate the TPR values at 1\% FPR. 
We want to point out that, although the ROC curves in~\autoref{fig:roc} already show the TPR for each possible operating point (i.e., the value of FPR), we report the results in~\autoref{tab:tpr_table} at 1\% FPR because it is a reasonable value commonly adopted in the literature~\cite{corona2017deltaphish,Demontis2017YesML}. We detail hereafter the key findings of our evaluations of \modsecurity against both the test set (\texttt{test}) and the adversarial test set (\texttt{test-adv}).

\myparagraph{Evaluation on Clean Samples.}
We first test \modsecurity on the data we have gathered, and the results of this first evaluation are indicated with red solid lines in~\autoref{fig:roc}. 
The ROC curve of PL1 (default \gls{PL} for \modsecurity) shows its inability to discriminate between benign and malicious SQL queries, with a TPR of 63.85\% at a 1\% \gls{FPR}.
The results for PL2 are the best among all \glspl{PL}, with a TPR of 66.82\% at 1\% FPR. The ROC curves for both PL3 and PL4 are almost identical, as PL4 has only two active rules more than PL3, which do not even improve its detection rate.

\myparagraph{Robustness against Adversarial SQLi.}
The results on the adversarial test set are indicated with red dashed lines in~\autoref{fig:roc}. The outcomes highlight a more alarming trend, thus clearly showing that \modsecurity is not able to withstand adversarial attacks.
Both datasets exhibited a pattern similar to the evaluation on the test set (\texttt{test}). Specifically, with the \wafamole dataset, the \gls{TPR} drops below 50\% at a 1\% \gls{FPR}, which is worse than random guessing. For the \modseclearn dataset, the situation is slightly better, but still concerning. Particularly with PL3 and PL4, the \gls{TPR} at 1\% \gls{FPR} is around 58\%, only slightly better than random guessing. This emphasizes that increasing the number of rules does not improve detection capabilities; instead, it worsens them by increasing false positives.

\begin{table*}[t]
\caption{TPR at 1\% FPR evaluated on the baseline/adversarial (\texttt{test}/\texttt{test-adv}) test sets for \modsecurity (ModSec);  \modseclearn (\gls{SVM}, \gls{LR}, \gls{SecSVM}, and \gls{RF}); and \advms (\gls{SVM}, \gls{LR}, \gls{SecSVM}, and \gls{RF}) using PL4. We also report the number of active rules (AR) for each model. Results are shown for both datasets: \wafamole and \modseclearn.}
\centering
\setlength{\tabcolsep}{5.5pt}
\begin{tabular*}{\textwidth}{@{\extracolsep{\fill}} l c c c c c c c c }
    \toprule
    & & PL1 & PL2 & PL3 & \multicolumn{4}{c}{PL4} \\
    \cmidrule(lr){6-9} & & & & & Base & AR & \advms & AR \\
    
    \midrule
    \multicolumn{9}{c}{\centering \textbf{\wafamole dataset}} \\
    \midrule
    
    \multirow{2}{*}{\centering ModSec vanilla} & test & 63.85\% & \textbf{66.82\%} & 61.00\% & 61.00\% & 62/62 & --- & \\ 
    & test-adv & 34.25\% & \textbf{45.86\%} & 39.12\% & 39.12\% & 62/62 & --- & \\
    
    \hline
    
    \multirow{2}{*}{\centering \modseclearn SVM ($\ell_1$)} & test & 69.40\% & 83.00\% & \textbf{86.84\%} & 86.80\% & 37/62 & 85.30\% & 42/62 \\ 
    & test-adv & 37.10\% & 59.90\% & 62.72\% & 63.97\% & 37/62 & \textbf{82.55\%} & 42/62 \\
    
    \hline
    
    \multirow{2}{*}{\centering \modseclearn SVM ($\ell_2$)} & test & 69.50\% & 83.10\% & 86.80\% & \textbf{86.80\%} & 50/62 & 85.25\% & 51/62 \\
    & test-adv & 39.20\% & 59.90\% & 62.71\% & 62.95\% & 50/62 & \textbf{81.40\%} & 51/62 \\
    
    \hline
    
    \multirow{2}{*}{\centering \modseclearn LR ($\ell_1$)} & test & 69.40\% & 83.00\% & \textbf{86.64\%} & 86.60\% & 29/62 & 84.10\% & 32/62 \\
    & test-adv & 36.50\% & 54.20\% & 60.91\% & 60.62\% & 29/62 & \textbf{79.85\%} & 32/62 \\
    
    \hline
    
    \multirow{2}{*}{\centering \modseclearn LR ($\ell_2$)} & test & 69.50\% & 82.54\% & 85.95\% & \textbf{85.95\%} & 50/62 & 84.05\% & 52/62\\
    & test-adv & 38.60\% & 53.75\% & 57.95\% & 58.37\% & 50/62 & \textbf{80.15\%} & 52/62 \\
    
    \hline
    
    \multirow{2}{*}{\centering \modseclearn RF} & test & 69.50\% & 83.89\% & 87.00\% & 87.00\% & 46/62 & \textbf{87.22\%} & 47/62 \\
    & test-adv & 41.35\% & 66.75\% & 64.5\% & 64.30\% & 46/62 & \textbf{84.90\%} & 47/62 \\
    
    \hline
    
    \multirow{2}{*}{\centering \modseclearn SecSVM} & test & 63.70\% & 80.78\% & 82.75\% & 82.58\% & 61/62 & \textbf{84.15\%} & 61/62 \\
    & test-adv & 57.65\% & 76.71\% & 61.96\% & 58.68\% & 61/62 & \textbf{81.84\%} & 61/62\\
    
    \midrule
    \multicolumn{9}{c}{\textbf{\modseclearn dataset}} \\
    \midrule
    
    \multirow{2}{*}{\centering ModSec vanilla} & test & \textbf{92.80\%} & 85.22\% & 75.71\% & 75.71\% & 62/62 & --- & \\
    & test-adv & \textbf{59.50\%} & 42.12\% & 31.07\% & 30.85\% & 62/62 & --- & \\
    
    \hline
    
    \multirow{2}{*}{\centering \modseclearn SVM ($\ell_1$)} & test & 92.80\% & \textbf{99.46\%} & 99.31\% & 99.41\% & 35/62 & 99.26\% & 41/62 \\
    & test-adv & 69.90\% & 70.87\% & 67.02\% & 69.86\% & 35/62 & \textbf{93.46\%} & 41/62 \\
    
    \hline
    
    \multirow{2}{*}{\centering \modseclearn SVM ($\ell_2$)} & test & 92.80\% & \textbf{99.46\%} & 99.31\% & 99.41\% & 49/62 & 99.26\% & 50/62 \\
    & test-adv & 69.80\% & 73.87\% & 71.47\% & 70.66\% & 49/62 & \textbf{92.86\%} & 50/62 \\
    
    \hline
    
    \multirow{2}{*}{\centering \modseclearn LR ($\ell_1$)} & test & 92.80\% & 99.46\% & 99.46\% & 99.46\% & 31/62 & \textbf{99.61\%} & 30/62 \\
    & test-adv & 73.05\% & 89.68\% & 89.95\% & 88.89\% & 31/62 & \textbf{94.03\%} & 30/62 \\
    
    \hline
    
    \multirow{2}{*}{\centering \modseclearn LR ($\ell_2$)} & test & 92.80\% & 99.46\% & 99.46\% & 99.46\% & 49/62 & \textbf{99.61\%} & 50/62 \\
    & test-adv & 73.25\% & 87.75\% & 86.92\% & 87.61\% & 49/62 & \textbf{94.29\%} & 50/62 \\
    
    \hline
    
    \multirow{2}{*}{\centering \modseclearn RF} & test & 92.80\% & 99.56\% & 99.56\% & 99.56\% & 49/62 & \textbf{99.65}\% & 50/62 \\
    & test-adv & 70.02\% & 86.20\% & 87.70\% & 87.45\% & 49/62 & \textbf{95.07\%} & 50/62\\
    
    \hline
    
    \multirow{2}{*}{\centering \modseclearn SecSVM} & test & 92.80\% & 99.50\% & 99.50\% & 99.50\% & 56/62 & \textbf{99.65\%} & 59/62\\
    & test-adv & 86.50\% & 97.76\% & 98.26\% & 98.26\% & 56/62 & \textbf{98.41\%} & 59/62\\
    \bottomrule
\end{tabular*}
\smallskip
\label{tab:tpr_table}
\end{table*}

\subsection{Evaluation of \mlms}
\label{sec:exp-ml}
We analyze the performance of \gls{SVM} and \gls{LR}, using $\ell_1$ and $\ell_2$ regularizations, \gls{SecSVM} and \gls{RF} \mlms against both the baseline clean and the adversarial samples.

\myparagraph{Evaluation on Clean Samples.}
We plot the ROC curves in~\autoref{fig:roc} using and green (\gls{RF}), violet (\gls{SecSVM}), blue (\gls{SVM} with $\ell_1$), light blue (\gls{SVM} with $\ell_2$), orange (\gls{LR} with $\ell_1$), brown (\gls{LR} with $\ell_2$), solid lines.
They clearly show the superiority of \mlms w.r.t. the respective \modsecurity counterpart regardless of the operating point, \ie, for any FPR value, the TPR of \mlms approaches is higher for all \glspl{PL} greater than 1. It is worth noting that, for PL1, the trained \gls{ML} models achieve similar results to the vanilla \modsecurity. This confirms that, even by learning optimal weights, rules enabled by PL1 are inappropriate to effectively discriminate benign samples from malicious ones. Finally, unlike the vanilla \modsecurity, all \mlms models achieve the best TPR for PL4 (even though the results for PL4 are slightly higher than those obtained for PL2). This result shows that, despite adding rules that may increase the FPR, machine learning can adjust their importance to improve the TPR/FPR trade-off. 

\myparagraph{Robustness against Adversarial \gls{SQLi}.}
As shown in~\autoref{fig:roc}, the \mlms models suffer the presence of adversarial attacks, particularly with the \wafamole dataset, but they still outperform the vanilla \modsecurity. Analyzing the results from the \wafamole dataset, it is evident that, the best performance is achieved with PL4, except for the \gls{RF} and \gls{SecSVM} models. This may be since, in this case, the adversarial \gls{SQLi} attacks are able to exploit the rules enabled by PL3 and PL4 to evade the model, by removing patterns that were considered important at training time. On the other hand, with the \modseclearn dataset, it is observed that starting from PL2, the detection capabilities exhibit a fairly consistent trend. While there is a slight deterioration, performance remains relatively stable across the different \glspl{PL}. Among all evaluated models, \gls{SecSVM} is the most robust, offering strong generalization and adversarial robustness. Even if \gls{RF} was included to explore non linear alternatives, its performance is comparable to linear models like \gls{SecSVM}, while worsening robustness. This confirms that linear models are sufficient to achieve excellent accuracy and are preferable given their robustness and transparency. In addition, linear models can be readily applied to the existing \gls{CRS} system by updating the rule weights after model training.\footnote{\url{https://owasp.org/www-project-waf-advanced-ruleset-management/}} This eases practical deployment while preserving
interpretability of decisions --- an important desideratum for web security.

\myparagraph{Imposing Sparsity through Regularization.}
We now analyze the effects of regularization by examining whether it is possible to select a reduced set of \gls{CRS} rules as features for classification. We employ an $\ell_1$ regularization term to impose sparsity on the trained models and assess its impact on the importance of each \gls{CRS} rule in the classification process. Additionally, we compare these results with those obtained using $\ell_2$ regularization, the default norm used by \gls{SVM} and \gls{LR}.
\begin{figure*}[h]
\centering
\includegraphics[width=0.95\textwidth]{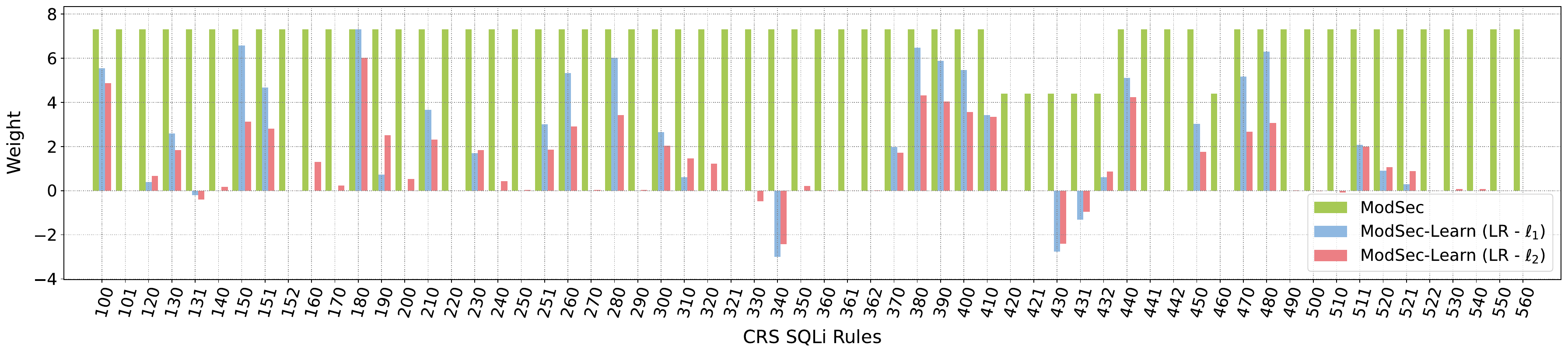}
\caption{Weight values learned at PL4 by \mlms \gls{LR}-$\ell_1$ (blue) and \mlms \gls{LR}-$\ell_2$ (light red), and the weights used by \modsecurity (green). Rules are expressed as the last three digits of their IDs (all starting with 942).}
     \label{fig:lr_weights}
\end{figure*}
\autoref{fig:lr_weights} displays the distribution of rule weights for \modseclearn implemented with \gls{LR} at PL4. We selected this \gls{PL} to activate all \gls{CRS} rules, providing a comprehensive overview of their impact. The blue and light red bars represent the weights calculated with $\ell_1$ and $\ell_2$ regularization, respectively, while the green bars represent the \modsecurity severity scores. Since the severity score ranges from 2 to 5, we normalized it using the minimum and maximum values of the \gls{LR} weights.
The results presented in~\autoref{tab:tpr_table} demonstrate that \gls{SVM} and \gls{LR} with $\ell_1$ regularization can achieve the same performance as the counterpart with $\ell_2$ regularization while utilizing fewer rules. Specifically, in the \wafamole dataset, the \gls{SVM} model employed 13 rules fewer than with the $\ell_2$ norm, and the \gls{LR} model 21 fewer than \gls{LR} with $\ell_2$. For the \modseclearn dataset, \gls{SVM} used 14 fewer rules than \gls{SVM} with $\ell_2$ norm, while \gls{LR} used 21 fewer.
Additionally, it is important to note that, compared to the 62 total rules in \gls{CRS}, linear models with $\ell_2$ regularization, as well as \gls{RF}, already reduce the number of rules used in the classification process compared to the vanilla \modsecurity. On average, these models use a maximum of 50 rules, effectively eliminating 12 rules deemed unnecessary for classification.
The rules assigned a weight of 0 by the \gls{ML} models are considered unnecessary for the classification task. Moreover, some rules may receive negative weights, suggesting that their presence might be more indicative of legitimate behavior rather than malicious activity.
Applying this approach to the \gls{CRS} introduces a more data-driven and less arbitrary method for selecting detection rules. Rather than rely on manual selection or a predefined set of rules that may not be optimal for the specific data being classified, \modseclearn enables automation of both rule selection and weight assignment, optimizing \gls{CRS}'s performance on the data.

\subsection{Evaluation of \advms}
\label{sec:exp-adv} 
\begin{figure}[!h]
    \centering
    \includegraphics[width=0.49\textwidth]{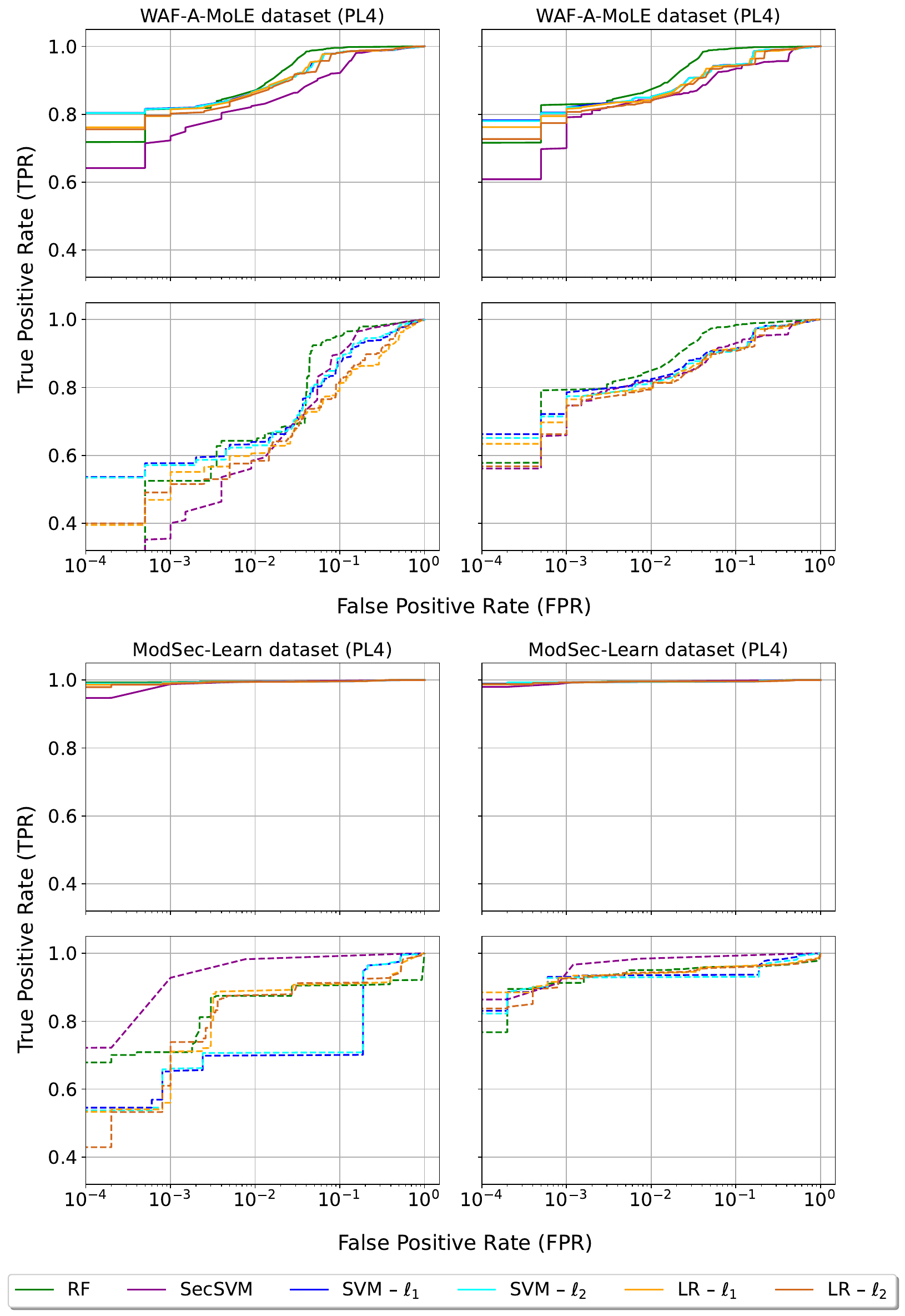}
     \caption{ROC curves of ModSec-Learn/AdvLearn (\textit{first/second column}) on \texttt{test}/\texttt{test-adv} (\textit{solid/dashed lines}), for \wafamole/\mlms  (\textit{top/bottom}) datasets.
     }
\label{fig:adv_train_results}
\end{figure}

Given the best results on PL4 among all the \glspl{PL} in terms of TPR/FPR, we select this configuration for re-training all \mlms models. We then evaluate the \advms against the \texttt{test} and \texttt{test-adv} sets of both datasets, and plot the results in~\autoref{fig:adv_train_results}. 
Also, we report the TPR of \advms at 1\% FPR in the second-to-last column of~\autoref{tab:tpr_table}.
Hereafter, we discuss the performance of \advms in comparison with \mlms. Overall, we observe that the robustness achieved by \advms clearly outperforms its non-hardened counterparts, \ie, \mlms.
Finally, we analyze the weights and predictions of \advms, and we thoroughly explain its remarkable level of adversarial robustness.

\myparagraph{Evaluation on Clean Samples.}
In the absence of attack, \advms has comparable performance to \mlms, except for SecSVM which shows an improvement in TPR on the \modseclearn dataset (cf. the violet solid lines in \autoref{fig:adv_train_results}, \textit{third row}). The reason is that SecSVM is retrained on less pessimistic attacks when considering problem-space AT, thereby yielding an improved robustness-accuracy tradeoff. 

\myparagraph{Robustness against Adversarial SQLi.}
Over the adversarial test set, \advms outperforms its non-hardened counterparts, especially when evaluated with the \wafamole dataset, reaching thus higher robustness (see the dashed lines of~\autoref{fig:adv_train_results}). Looking at the results in detail, with the \wafamole dataset, the re-trained models improve the average detection performance by 33\% compared to their non-hardened versions. For the \modseclearn dataset, the improvement is less marked in some models, such as \advms \gls{LR}, but it is still present. Moreover, considering the \wafamole dataset for example, the best \advms (\ie, \gls{RF} at PL4) is 85\% more robust than the best vanilla \modsecurity (PL2).
Of course \advms is still vulnerable to new adversarial examples optimized against it, but the decrement in performance is lower compared to the decrement caused by the non-hardened models.

\begin{figure*}[!h]
\centering
\includegraphics[width=0.95\textwidth]{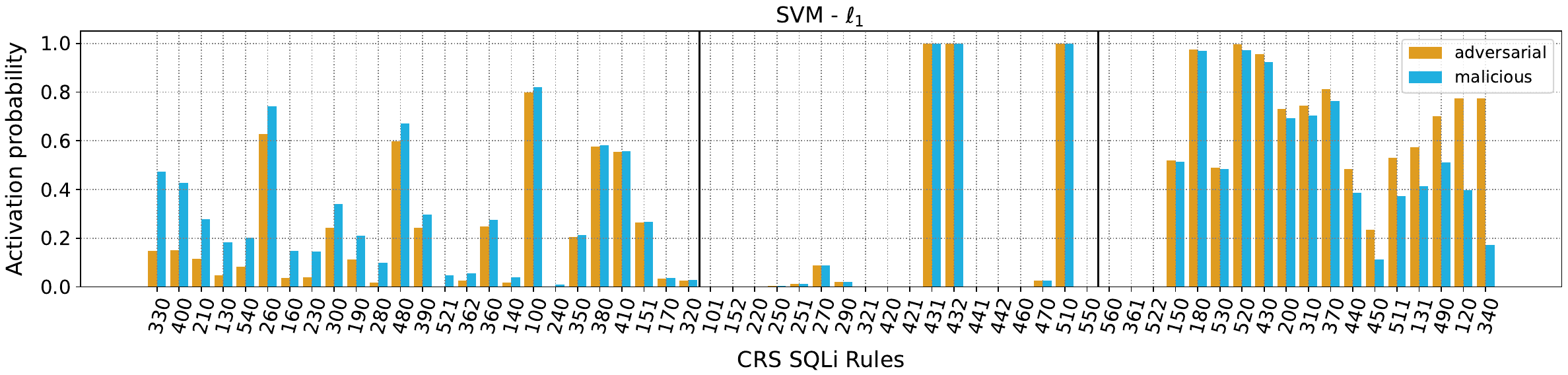}
\includegraphics[width=0.95\textwidth]{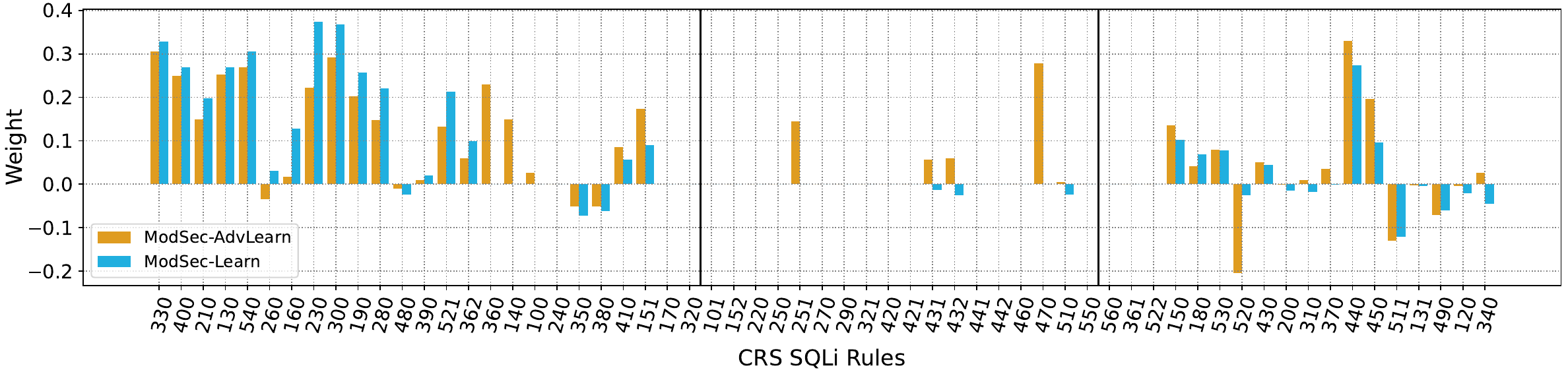}
\caption{ \textit{Top:} Activation probability of \gls{CRS} rules (expressed using the last three digits of their IDs, which always starts with 942) on malicious/adversarial (cyan/orange) \gls{SQLi} samples optimized against \modseclearn \gls{SVM}-$\ell_1$ on the \wafamole dataset. 
\textit{Bottom:} Rule weights learned by \modseclearn/\advms \gls{SVM}-$\ell_1$ (cyan/orange) on the \wafamole dataset.
}
\label{fig:rules_weigths_activation_prob}
\end{figure*}

\myparagraph{Explaining Robustness of \advms.}
Here we explain why \advms achieves better robustness, focusing on the linear SVM model. 
First, we compute the \emph{rule activation delta} as $\Delta \vct a_i = \vct a_i - \vct a^\prime_i$. It captures the difference between the probability that rule \emph{i} is activated by standard SQLi attacks ($\vct a_i$) and by their adversarial counterparts ($\vct a^\prime_i$).
If $\Delta \vct a_i$ is positive (negative), it means that \wafamole is bypassing (activating) rule \emph{i}, and if it is zero it means that \wafamole does not affect rule \emph{i}.
Second, as we are considering a linear model, we also inspect its feature weights and observe how they change when the same model is retrained on adversarial \gls{SQLi} queries.
Within this scenario, we analyze how each rule is affected by the adversarial \gls{SQLi} attacks generated through \wafamole (\autoref{fig:rules_weigths_activation_prob}), as well as how differently the baseline \mlms SVM and the \advms SVM compute weights for each rule (\autoref{fig:rules_weigths_activation_prob}).
In \autoref{fig:rules_weigths_activation_prob} we plot the probability that a rule is active, and we display how different the distributions induced by the malicious (cyan) and adversarial (yellow) SQLi queries are. The rules are sorted by their value of $\Delta \vct a_i$, and grouped into three classes (separated by vertical black lines): rules evaded by \wafamole (left), rules that \wafamole is unable to bypass (center), and rules that are triggered only by adversarial attacks as a side effect (right).
The same order is also used for the weights of the \mlms \gls{SVM} and the \advms \gls{SVM} shown in~\autoref{fig:rules_weigths_activation_prob}.
We can observe that more than one-third of the rules are exploited by \wafamole to avoid detection, as the first group has a drop in the probability of being active.
This is also confirmed by~\autoref{fig:rules_weigths_activation_prob}, where we can see that most of the positive weights (i.e., the ones that increase the scores towards the malicious class) assigned by \mlms (cyan) are all concentrated in the first group, which is exactly the one leveraged by adversarial attacks.
Conversely, \advms (yellow) is more robust since it spreads the importance on more rules, prioritizing the ones belonging to the second and third groups, making attacks harder to land and easier to detect.
Of course, in this analysis, we focus on a linear model, which, as shown in~\autoref{sec:exp-adv} is still vulnerable to adversarial attacks.
Indeed, by looking at~\autoref{fig:rules_weigths_activation_prob}, \advms attributes negative weights (i.e., those that decrease the score towards the benign class) to some features of the first block, hence leaving the ability to \wafamole to find some successful adversarial \gls{SQLi} queries.
On the other hand, as shown in~\autoref{fig:adv_train_results}, for the most of the cases \advms \gls{RF} is more robust than the linear models counterpart as its non-linearity exploits relationships among different rules, forcing the attacker to manipulate more rules in a consistent manner to bypass detection.

\section{Related work}\label{sec:related_work}

In this section, we briefly review related work analyzing the performance and robustness of ModSecurity and the CRS, and conclude by discussing the main differences of our current work with our preliminary results in~\cite{scano2024modsec}.

\myparagraph{ModSecurity and CRS.} 
Previous work has considered the impact of different types of web security threats on ModSecurity when using the CRS configurations~\cite{singh2018_ms_pl_study,sobola_modsecurity_study}. However, unlike ours, a very limited number of attack samples is normally used (e.g., \cite{singh2018_ms_pl_study} uses only 27 samples), without providing any detailed investigation of the trade-off between TPR and FPR. Other approaches~\cite{betarte2018web, montes2021web} have applied \gls{ML} to detect web threats and only used \modsecurity as a baseline for comparison, without even evaluating adversarial robustness.
%Furthermore, no previous work has analyzed how the detection rate of the CRS configurations used in ModSecurity deteriorates under adversarial manipulations of the input queries. 

\myparagraph{Adversarial SQLi.} Other work has considered adversarial \gls*{SQLi} attacks against \modsecurity by proposing different approaches based on \gls*{ML} \cite{ml_driven}, \acrfull{RL} \cite{wang2020evading,hemmati2022rlwaf,guan2023ssqli}, fuzzing techniques \cite{demetrio2020waf,LiYV22}, and heuristic search algorithm like Monte-Carlo tree search \cite{autospear}.
However, the reported results are partial (\eg, \cite{autospear,hemmati2022rlwaf} just limit the analysis to the default PL1) and do not explain precisely why \modsecurity is failing and how it could be improved.
To the best of our knowledge, no prior research has conducted a comprehensive analysis of the CRS for ModSecurity as we have done in this work.
Additionally, no previous studies have explored the potential of adversarial training in this domain, making us the first to propose a robust ML methodology for effectively enhancing the robustness of WAFs. 

\myparagraph{ModSec-Learn.} With respect to our work introducing \mlms in~\cite{scano2024modsec}, we have extended here our approach as follows: 
(i) we have examined the impact of adversarial attacks on the CRS within the SQL domain; 
(ii) we have increased the robustness of our approach by developing a novel adversarial training procedure (\advms); 
(iii) we have investigated whether strongly-regularized models could withstand adversarial SQLi attacks, devising a novel version of SecSVM; (iv) we have analyzed how the baseline and \advms compute weights for each rule, highlighting which rules are more robust to perturbations; and (v) we have included an additional dataset~\cite{demetrio2020waf} in our experiments.
\section{Conclusions and Future Work}\label{sec:conclusions}
In this work, we proposed \advms, a novel methodology for training \gls{ML} classifiers using the \gls{CRS} rules as input features. This allows learning how to optimally tune the severity levels (\ie, the weights) of the \gls{CRS} rules,
yielding an improved trade-off between detection and false positive rates.
Furthermore, our approach relies upon a novel problem-space adversarial training procedure that incorporates knowledge of state-of-the-art \gls{SQLi} manipulations to counter the presence of adversarial \gls{SQLi} attacks.
Among the main findings, we show that \advms improves the detection rate of the vanilla \modsecurity by 30\%, while removing 50\% of the \gls{CRS} rules through embedded feature selection with $\ell_1$ regularization. It also improves adversarial robustness up to 85\% via robust \textit{linear} models, without hindering interpretability of decisions and providing ease of integration with the current \gls{CRS} implementations.
We can thus state that our methodology provides a first, concrete example of how adversarial machine learning can be used to effectively enhance the robustness of \glspl{WAF} against adversarial attacks, highlighting novel, promising directions towards designing robust machine learning models for cybersecurity-related applications.

We foresee several other promising avenues for advancing our work. First, although in this work we only target \gls{SQLi} attacks, our methodology is general enough to tackle other web threats like \gls{XSS}. In this direction, future work could also explore the integration of automated pentesting tools that leverage large language models~\cite{deng2024pentestgpt}, to extend the capabilities of \wafamole. Second, we also see future developments in evaluating other state-of-the-art ML-based \glspl{WAF}. Indeed, we think that the same results can also be obtained on more advanced models such as \gls{CNN}~\cite{luo2019cnn}, as well as on different feature representation approaches~\cite{sqligot, montes2021web}.
This is also true for commercial \glspl{WAF}. To this end, an interesting future extension of this work is to evaluate them in terms of transferability~\cite{demontis2019transferability} of adversarial \gls{SQLi} attacks optimized on \modsecurity.

\section*{Acknowledgments}
This research has been partly supported by the TESTABLE project, funded by the EU H2020 research and innovation program (grant no. 101019206); the ELSA project, funded by the Horizon Europe research and innovation program (grant no. 101070617); projects FAIR (PE00000013) and  SERICS (PE00000014) under the NRRP MUR program funded by the EU – NGEU. This work was carried out while C.~Scano was enrolled in the Italian National Doctorate on AI run by the Sapienza University of Rome in collaboration with the University of Cagliari.

% \section*{Acknowledgment}
% This research was supported by the TESTABLE project, funded by the European Union's Horizon 2020 research and innovation program (grant no. 101019206); the ELSA project, funded by the European Union's Horizon Europe research and innovation program (grant agreement no. 101070617); and the KINAITICS project, funded by the European Union's Horizon Europe research and innovation program (grant agreement no. 101070176).

\bibliographystyle{ieeetr}
\bibliography{bibliography}

% \input{src/appendix}
% Bisogna mettere qui tutte le informazioni degli autori (SHORT BIO)

\vskip -1\baselineskip plus -0fil

\begin{IEEEbiography}[{\includegraphics[width=1in,height=1.25in,clip,keepaspectratio]{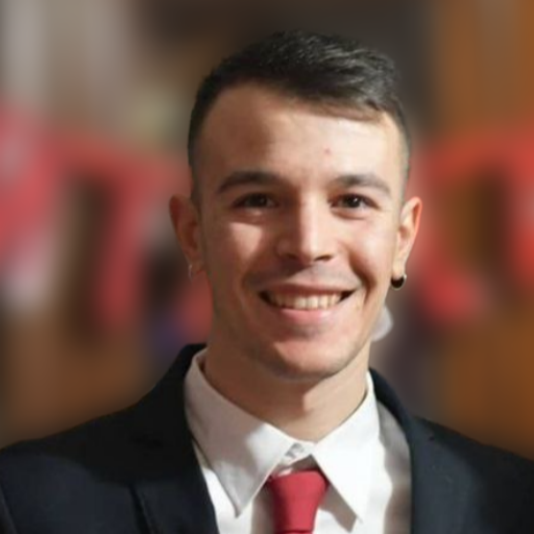}}]{Giuseppe Floris} received his BSc degree in Electrical, Eletronical and Computer Engineering in 2021, and his MSc degree in Computer Engineering, Cybersecurity, and Artificial Intelligence with honors in September 2023 from the University of Cagliari.
He is currently a Ph.D. student in electronics and computer engineering at the University of Cagliari.
His research focuses on Adversarial Machine Learning and its applications in the cybersecurity domain.
\end{IEEEbiography}

\vskip -1\baselineskip plus -1fil

\begin{IEEEbiography}[{\includegraphics[width=1in,height=1.25in,clip,keepaspectratio]{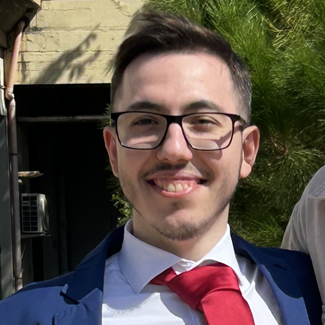}}]{Christian Scano} received his BSc degree in Computer Science in 2021 and his MSc degree in Computer Engineering, Cybersecurity, and Artificial Intelligence with honors in September 2024 from the University of Cagliari. He is currently enrolled in the national PhD in Artificial Intelligence and Computer Security at University of Cagliari and Sapienza University of Rome. His research focuses on Web Application Security and Machine Learning.
\end{IEEEbiography}

\vskip -1\baselineskip plus -1fil

\begin{IEEEbiography}[{\includegraphics[width=1in,height=1.25in,clip,keepaspectratio]{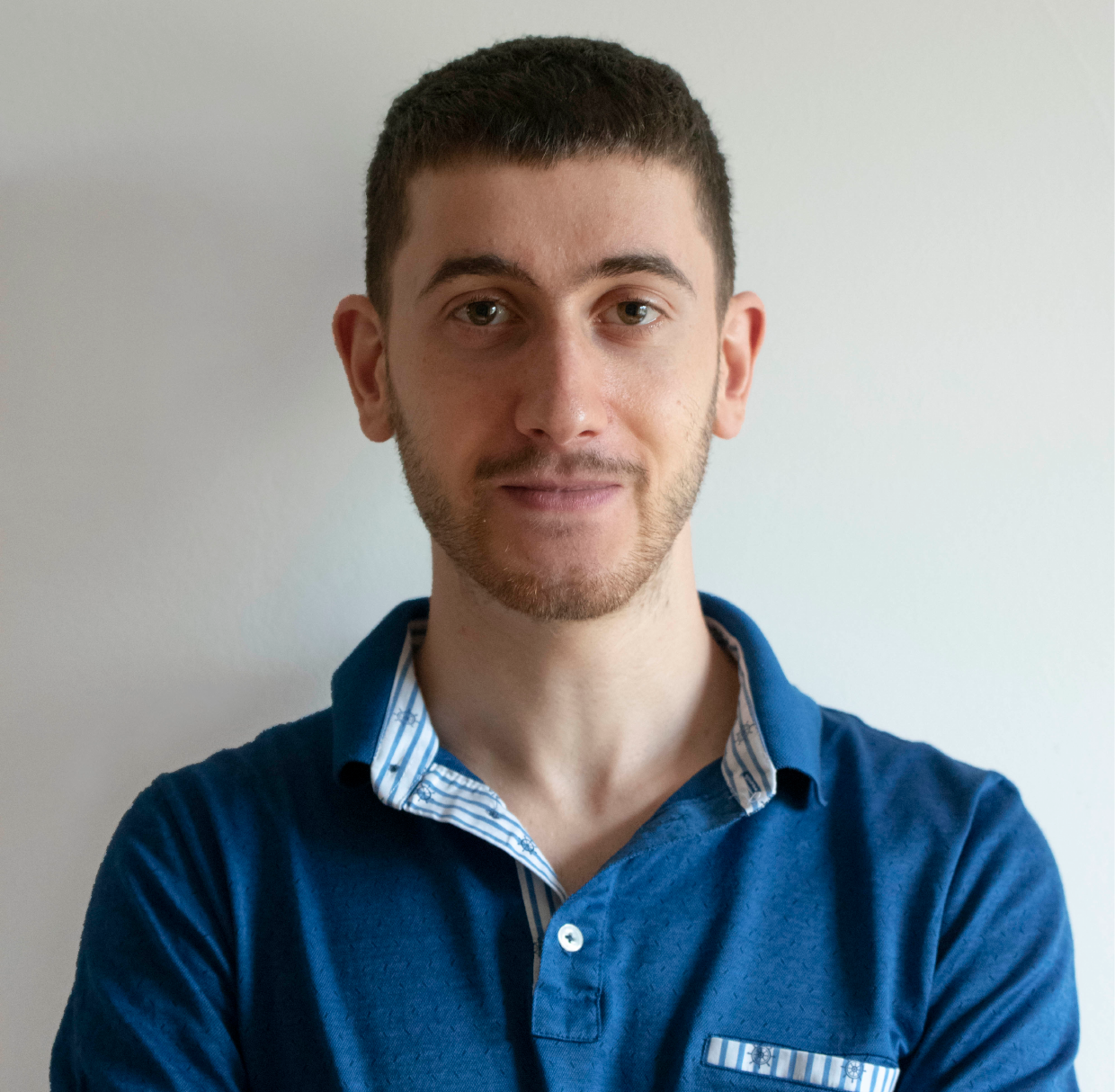}}]{Biagio Montaruli} received his B.Sc. and M.Sc. degrees in computer engineering from the Polytechnic University of Bari (Bari), in 2018 and 2021, respectively.
He is currently a Ph.D. candidate in artificial intelligence and computer security at EURECOM (France).
His research focuses on adversarial machine learning, with strong focus on its application in the cyber-security domain.
\end{IEEEbiography}

\vskip -1\baselineskip plus -1fil

\begin{IEEEbiography}[{\includegraphics[width=1in,height=1.25in,clip,keepaspectratio]{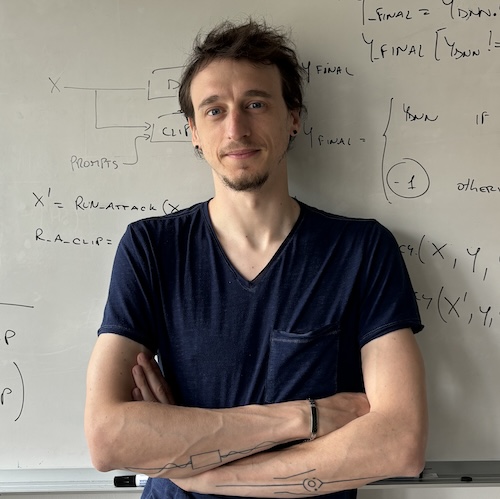}}]{Luca Demetrio} (MSc 2017, PhD 2021) is an Assistant Professor at the University of Genoa, investigating the security of Windows malware detectors implemented with Machine Learning techniques. He is part of the development team of SecML, and the maintainer of SecML Malware, a Python library for creating adversarial Windows malware.
\end{IEEEbiography}

\vskip -1\baselineskip plus -1fil

\begin{IEEEbiography}[{\includegraphics[width=1in,height=1.25in,clip,keepaspectratio]{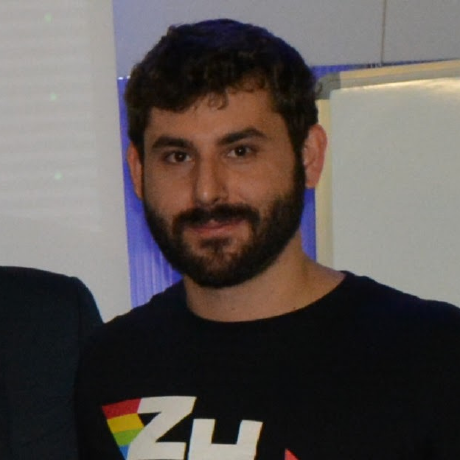}}]{Andrea Valenza} is an Application Security Engineer at Prima Assicurazioni.
He received his PhD from the University of Genova, with a thesis focused on novel vulnerabilities, including counter-attacking automatic security scanners.
His current research interest is automated security testing, with a focus on improving (and bypassing) regex-based validation, and detection of vulnerabilities via static analysis.
\end{IEEEbiography}

\vskip -1\baselineskip plus -1fil

\begin{IEEEbiography}[{\includegraphics[width=1in,height=1.25in,clip,keepaspectratio]{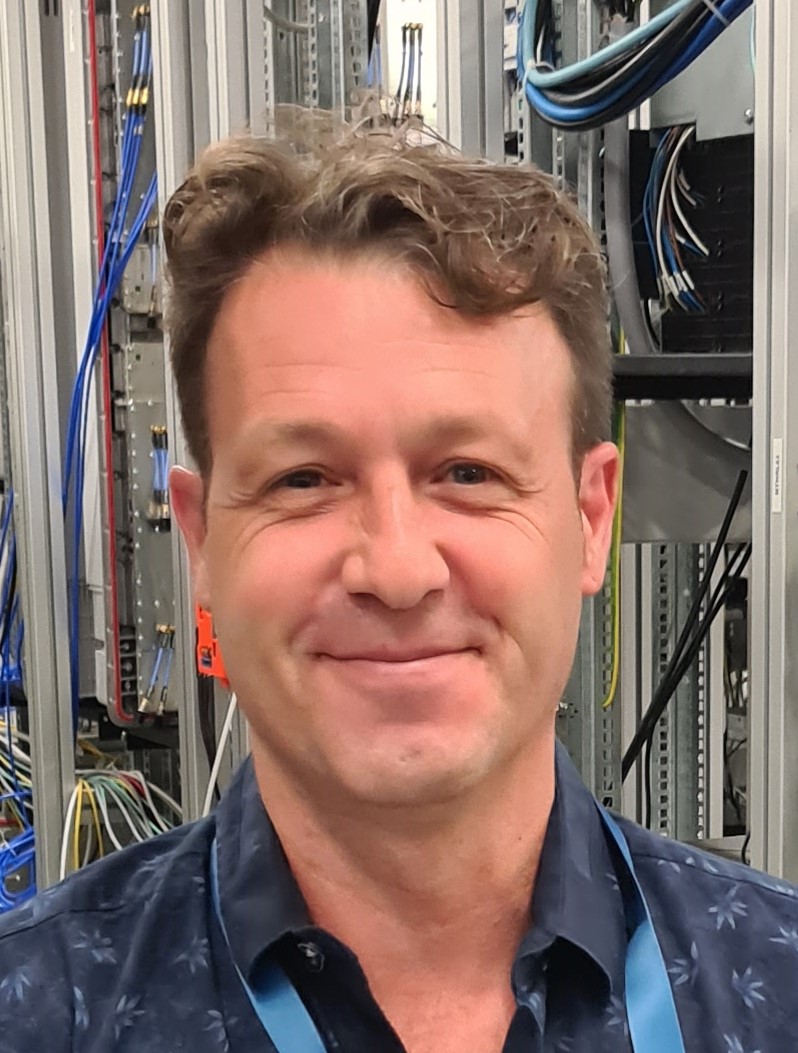}}]{Luca Compagna} works at Endor Labs, contributing to the software security analysis research area. He received his Ph.D. in Computer Science jointly from the U. of Genova and U. of Edinburgh, working on security protocols analysis. His areas of interests include security testing, security engineering, automated reasoning, and their application to the modeling and analysis of industrial relevant scenarios. After some work on DAST techniques for cross domain web-based scenarios and CSRF experiments, he recently focused his attention to static analysis and to the security testing of AI-based components.
\end{IEEEbiography}

\vskip -1\baselineskip plus -1fil

\begin{IEEEbiography}[{\includegraphics[width=1in,height=1.25in,clip,keepaspectratio]{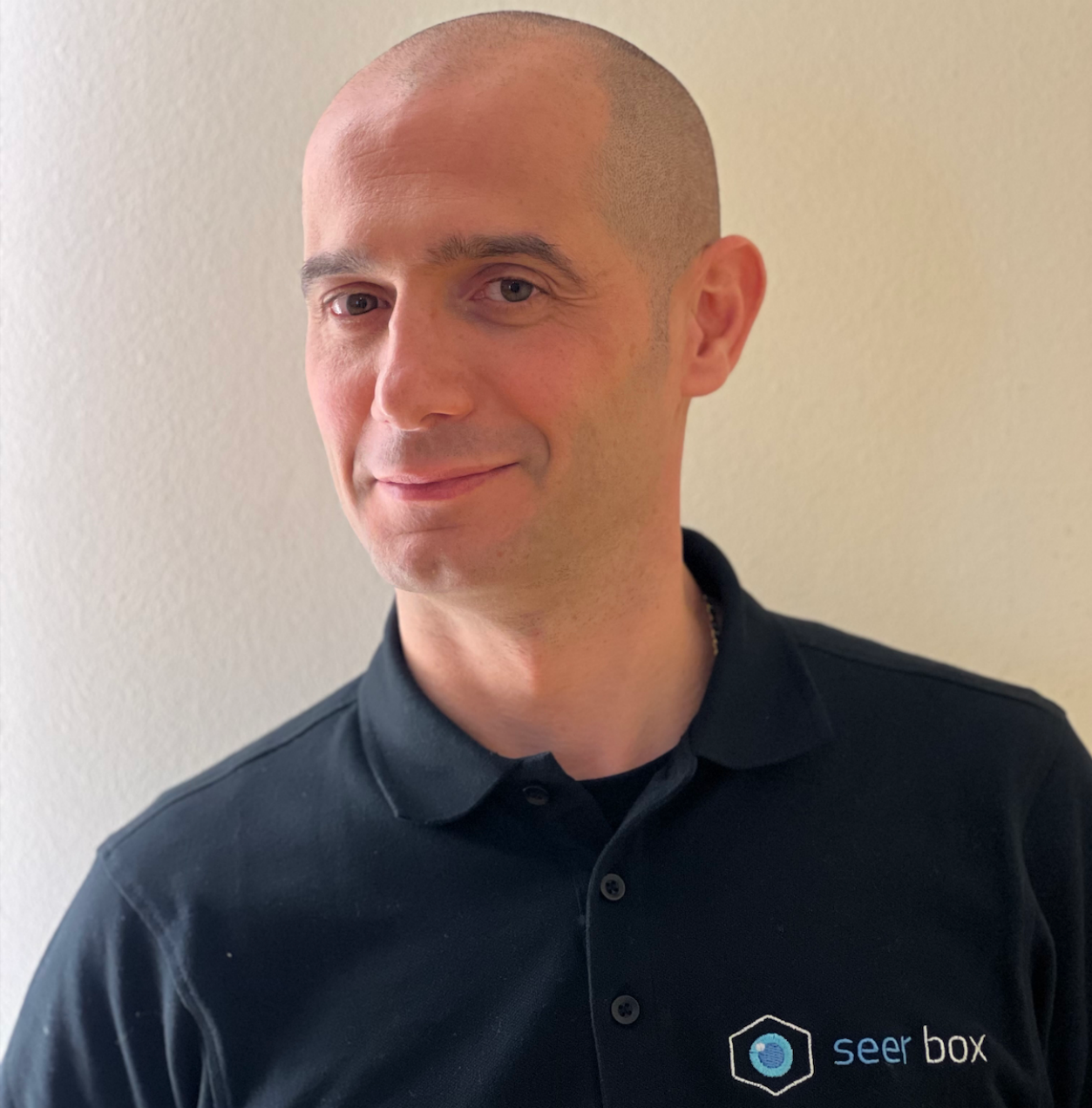}}]{Davide Ariu} received a PhD in electronics and computer engineering from the University of Cagliari. He is co-founder and CEO of Pluribus One (http://www.pluribus-one.it) and co-chair of the OWASP (http://owasp.org) Italy Chapter.
\end{IEEEbiography}

\vskip -1\baselineskip plus -1fil

\begin{IEEEbiography}[{\includegraphics[width=1in,height=1.25in,clip,keepaspectratio]{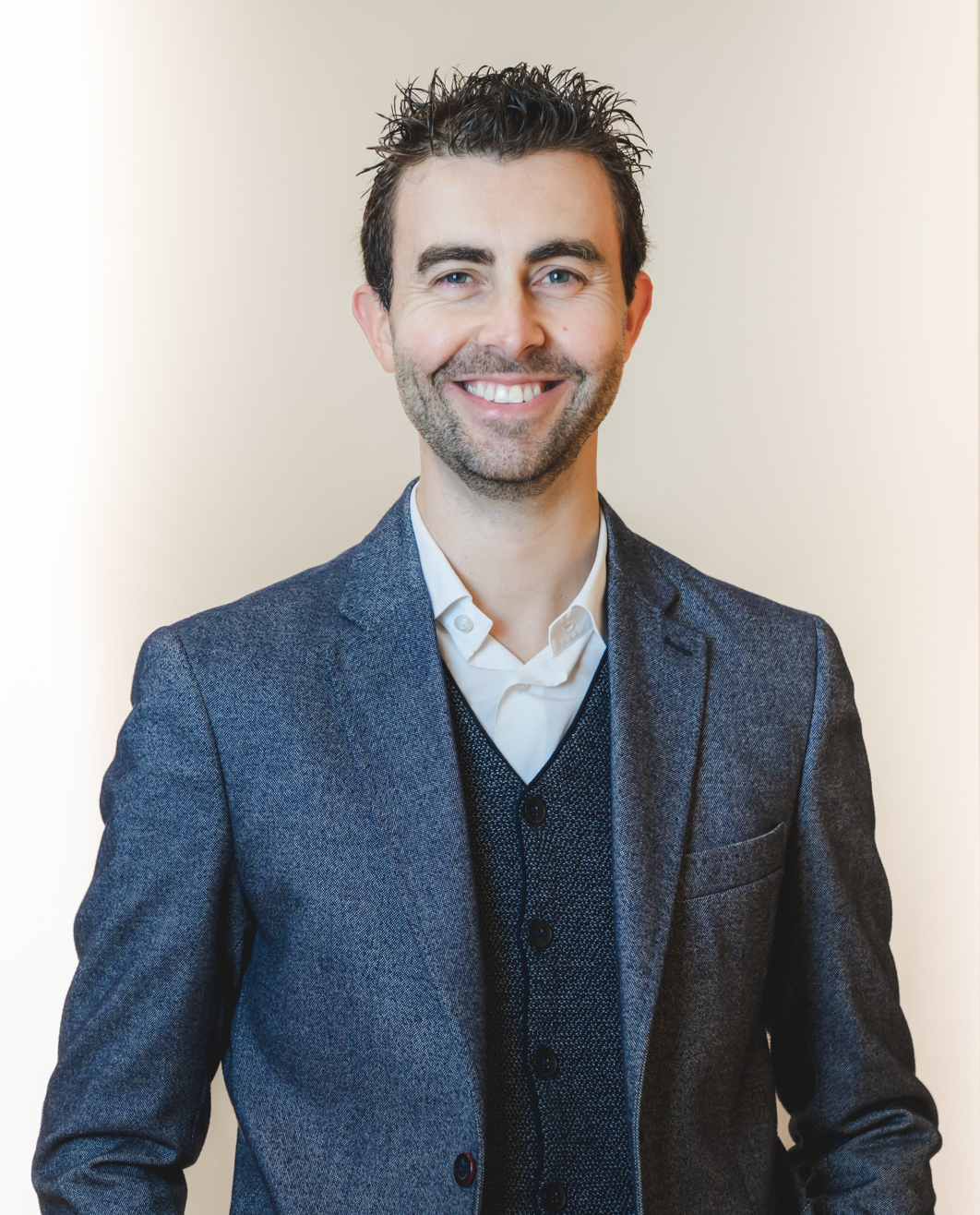}}]{Luca Piras} is the Operation Manager and Co-founder of Pluribus One. He received his MSc Degree in Electronic Engineering in 2007 and his Doctor Europaeus and PhD Degree in Computer Engineering in 2011, both from the University of Cagliari. His expertise lies in Computer Vision, Pattern Recognition, and Machine Learning. His research has been published in several international, peer-reviewed journals and conferences. At Pluribus One, he is responsible for several EU-funded R\&D projects and is a member of the OWASP Foundation.
\end{IEEEbiography}

\vskip -1\baselineskip plus -1fil

\begin{IEEEbiography}[{\includegraphics[width=1in,height=1.25in,clip,keepaspectratio]{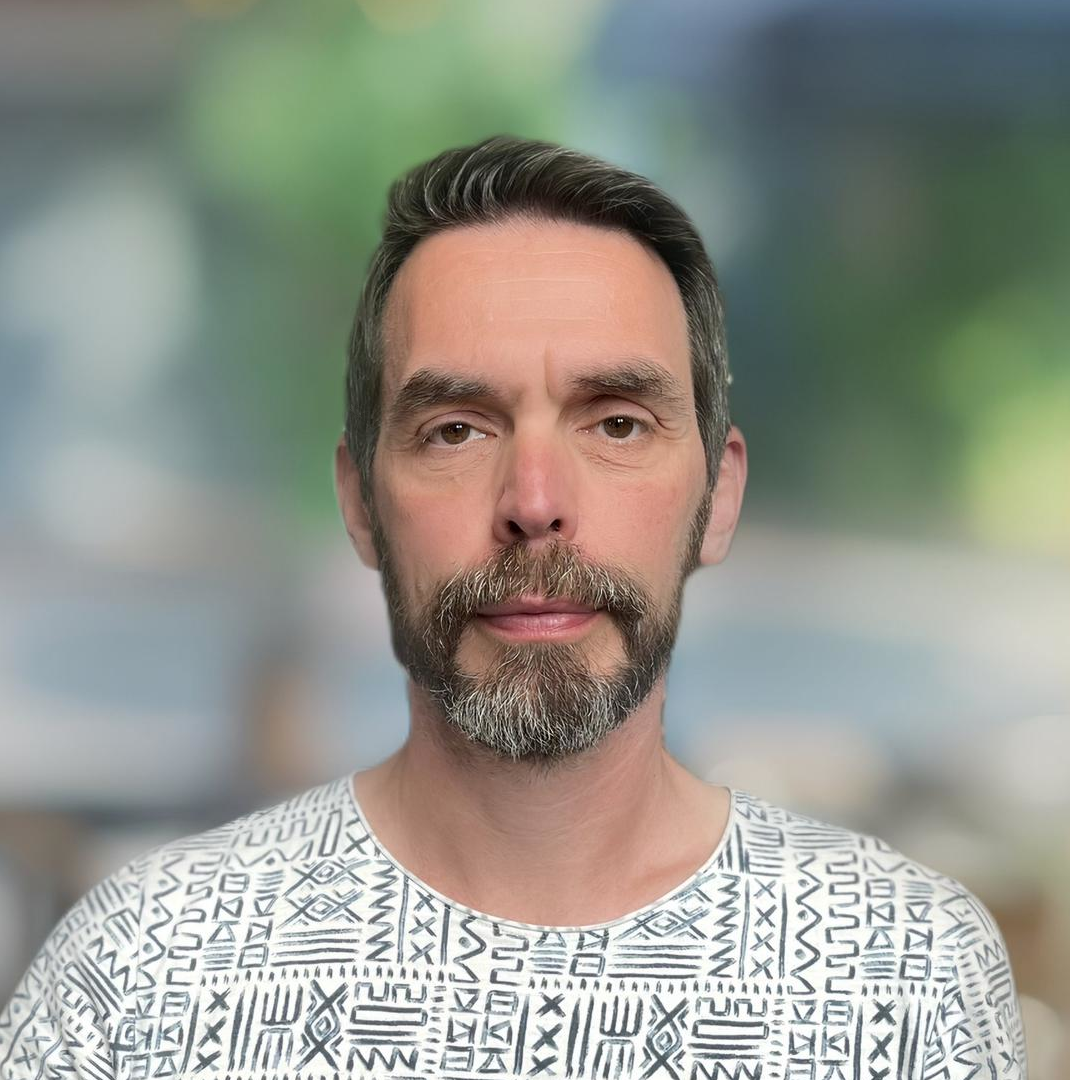}}]{Davide Balzarotti} is a full Professor and head of the Digital Security Department at EURECOM. His research interests include most aspects of software and system security and in particular the areas of binary and malware analysis, reverse engineering, computer forensics, and web security.
Davide authored more than 100 publications in leading conferences and journals. He has been the Program co-Chair of Usenix Security 2024, ACSAC 2017, RAID 2012, and Eurosec 2014.
He received an ERC Consolidator and an ERC PoC Grant for his research in the analysis of compromised systems.
Davide is also a member of the ``Order of the Overflow'' team, which organized the DEF CON CTF competition between 2018 and 2021.
\end{IEEEbiography}

\vskip -1\baselineskip plus -1fil

\begin{IEEEbiography}[{\includegraphics[width=1in,height=1.25in,clip,keepaspectratio]{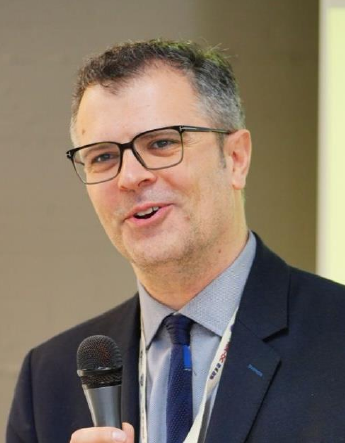}}]{Battista Biggio} (MSc 2006, PhD 2010) is Full Professor of Computer Engineering at the University of Cagliari, Italy. He has provided pioneering contributions to machine learning security. His paper ``Poisoning Attacks against Support Vector Machines'' won the prestigious 2022 ICML Test of Time Award.  He chaired IAPR TC1 (2016-2020), and served as Associate Editor for IEEE TNNLS and IEEE CIM. He is now Associate Editor-in-Chief for Pattern Recognition and serves as Area Chair for NeurIPS and IEEE Symp. SP. He is Fellow of IEEE and AAIA, ACM Senior Member, and Member of IAPR, AAAI, and ELLIS. 
\end{IEEEbiography}

\end{document}